%% file: main.tex
\documentclass[twoside,11pt]{article}
\usepackage{blindtext}

\input{shortcuts}

\usepackage{dmlr2e}

\usepackage{amssymb}
\usepackage{lastpage}
\dmlrheading{23}{2025}{1-\pageref{LastPage}}{12/20; Revised 04/14}{06/16}{21-0000}{Jin et al} 
\def\openreview{\url{https://openreview.net/forum?id=BJnusBahD3}} 

\ShortHeadings{The FIX Benchmark}{Jin et al.}
\firstpageno{1}

\begin{document}

\title{The FIX Benchmark: \\Extracting Features Interpretable to eXperts}


\newcommand{\penncis}{$^\text{\brachio}$}
\newcommand{\pennphysics}{$^\bigstar$}
\newcommand{\pennmed}{$^\dagger$}
\newcommand{\toronto}{$^\ddag$}

\author{\name Helen Jin\penncis $^*$ 
        \email helenjin@seas.upenn.edu\\
        \name Shreya Havaldar\penncis $^*$ 
        \email shreyah@seas.upenn.edu\\
        \name Chaehyeon Kim\penncis $^*$ 
        \email chaenyk@seas.upenn.edu\\
        \name Anton Xue\penncis $^*$ 
        \email antonxue@seas.upenn.edu\\
        \name Weiqiu You\penncis $^*$ 
        \email weiqiuy@seas.upenn.edu\\
        \name Helen Qu\pennphysics
        \email helenqu@sas.upenn.edu\\
        \name Marco Gatti\pennphysics
        \email mgatti29@sas.upenn.edu\\
        \name Daniel A. Hashimoto\pennmed
        \email daniel.hashimoto@pennmedicine.upenn.edu\\
        \name Bhuvnesh Jain\pennphysics
        \email bjain@physics.upenn.edu\\
        \name Amin Madani\toronto
        \email amin.madani@uhn.ca\\
        \name Masao Sako\pennphysics
        \email masao@sas.upenn.edu\\
        \name Lyle Ungar\penncis
        \email ungar@seas.upenn.edu\\
        \name Eric Wong\penncis
        \email exwong@seas.upenn.edu\\ \\
        \addr \penncis \textit{Department of Computer and Information Science, University of Pennsylvania, USA} \\
        \addr \pennphysics Department of Physics and Astronomy, University of Pennsylvania, USA \\
        \addr \pennmed Department of Surgery, Perelman School of Medicine, University of Pennsylvania, USA \\
        \addr \toronto Department of Surgery, University of Toronto, Canada \\
       }

\makeatletter
\renewcommand{\@makefntext}[1]{%
  \noindent\makebox[0.9em][l]{\@thefnmark}#1}
\makeatother

\def\thefootnote{*}\footnotetext{Equal contribution.}

\editor{Hugo Jair Escalante}

\maketitle

\input{sections/abstract}


\input{sections/introduction}

\input{sections/related_work}
\input{sections/problem_settings}

\input{sections/datasets}
\input{datasets/cosmology}
\input{datasets/supernova}

\input{datasets/politeness}

\input{datasets/emotion}

\input{datasets/chest_xray}

\input{datasets/abdomen}

\input{sections/baselines}
\input{sections/conclusion}

\input{sections/broader_impact}

\input{sections/acknowledgement}


\vskip 0.2in
\bibliography{main}

\appendix
\input{sections/appendix}

\end{document}

%% file: shortcuts.tex
\usepackage[utf8]{inputenc} 
\usepackage[T1]{fontenc}    
\usepackage{url}            
\usepackage{booktabs}       
\usepackage{amsfonts}       
\usepackage{nicefrac}       
\usepackage{microtype}      
\usepackage{xcolor}         
\usepackage{bbm}

\usepackage{graphicx}
\usepackage[caption=false]{subfig}

\usepackage{amsmath}
\usepackage{amsthm}
\usepackage{amssymb}

\usepackage{makecell}
\usepackage{tabularx}

\definecolor{prussian}{HTML}{006fd4}

\definecolor{skyblue}{HTML}{448EE4}

\usepackage{xspace}
\newcommand{\dataset}{\textsc{FIX}\xspace}
\newcommand{\datasetscore}{\textsc{FIXScore}\xspace}
\newcommand{\expertalign}{\textsc{ExpertAlign}\xspace}
\newcommand{\Om}{$\Omega_m$}
\newcommand{\seight}{$\sigma_{8}$}

\input{urls}

\usepackage{mathtools}
\DeclarePairedDelimiter\parens{(}{)}
\DeclarePairedDelimiter\bracks{[}{]}

\DeclarePairedDelimiter\abs{\lvert}{\rvert}
\DeclarePairedDelimiter\norm{\lVert}{\rVert}

\newcommand\mbb[1]{\mathbb{#1}}

\newcommand\mcal[1]{\mathcal{#1}}
\newcommand\mrm[1]{\mathrm{#1}}

\newcommand\msc[1]{\textsc{#1}}

\definecolor{myblue}{HTML}{0000ff}

\usepackage{amsmath}

\DeclareMathOperator*{\argmin}{arg\,min}

\usepackage{array, multicol, multirow}
\usepackage{enumitem}

\newcommand{\textexpertfeatfont}[1]{{\fontfamily{lmtt}\selectfont #1}}

\newcommand{\brachio}{\ensuremath{%
  \mathchoice{\includegraphics[height=2ex]{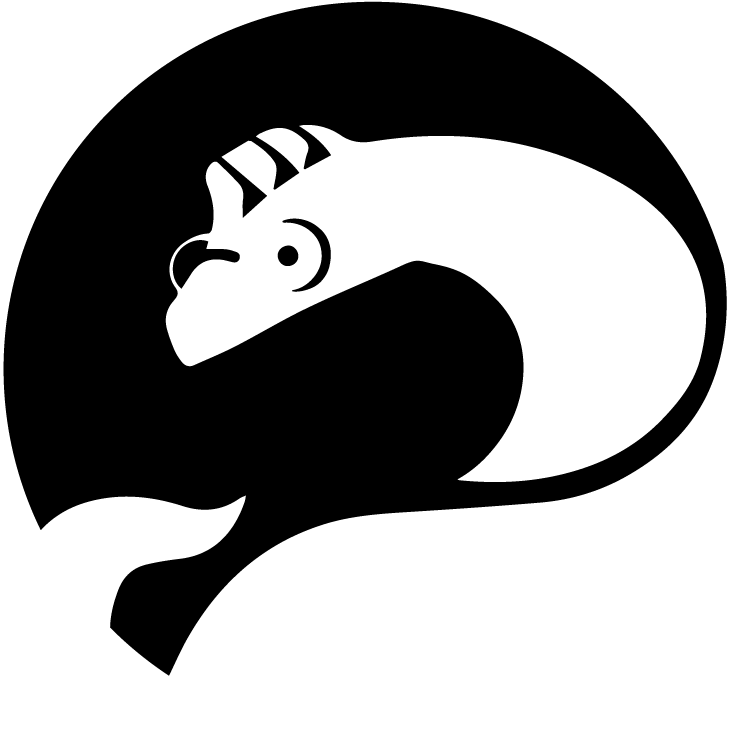}}
    {\includegraphics[height=2ex]{BrachioLabLogo_BW_NoText_Circle.png}}
    {\includegraphics[height=1.5ex]{BrachioLabLogo_BW_NoText_Circle.png}}
    {\includegraphics[height=1ex]{BrachioLabLogo_BW_NoText_Circle.png}}
}}

%% file: urls.tex
\newcommand{\massmapsurl}{https://huggingface.co/datasets/BrachioLab/massmaps-cosmogrid-100k}
\newcommand{\supernovaurl}{https://huggingface.co/datasets/BrachioLab/supernova-timeseries}
\newcommand{\politenessurl}{https://huggingface.co/datasets/BrachioLab/multilingual_politeness}
\newcommand{\emotionurl}{https://huggingface.co/datasets/BrachioLab/emotion}
\newcommand{\chestxurl}{https://huggingface.co/datasets/BrachioLab/chestx}
\newcommand{\cholecurl}{https://huggingface.co/datasets/BrachioLab/cholec}

%% file: sections/abstract.tex
\begin{abstract}
Feature-based methods are commonly used to explain model predictions, but these methods often implicitly assume that interpretable features are readily available.
However, this is often not the case for high-dimensional data, and it can be hard even for domain experts to mathematically specify which features are important.
Can we instead automatically extract collections or groups of features that are aligned with expert knowledge? To address this gap, we present FIX (Features Interpretable to eXperts), a benchmark for measuring how well a collection of features aligns with expert knowledge. 
In collaboration with domain experts, we propose FIXScore, a unified expert alignment measure applicable to diverse real-world settings across cosmology, psychology, and medicine domains in vision, language, and time series data modalities.
With FIXScore, we find that popular feature-based explanation methods have poor alignment with expert-specified knowledge, highlighting the need for new methods that can better identify features interpretable to experts. 
\end{abstract}

%% file: sections/introduction.tex
\section{Introduction}
\label{sec:intro}
Machine learning is increasingly used in domains like healthcare~\citep{DBLP:journals/corr/abs-1907-07374}, law~\citep{ATKINSON2020103387}, governance~\citep{doi:10.1080/01900692.2019.1575664}, science~\citep{de_la_Torre_L_pez_2023}, education~\citep{10.1007/978-3-319-93843-1_12} and finance~\citep{modarres2018towards}.
However, modern models are often black-box, which makes it hard for practitioners to understand their decision-making and safely use model outputs~\citep{rai2019explanation}.
For example, surgeons are concerned that blind trust in model predictions will lead to poorer patient outcomes~\citep{Hameed2023what};
in law, there are known instances of wrongful incarcerations due to over-reliance on faulty model predictions~\citep{Zeng_2016,wexler2017computer}.
Although such models have promising applications, their opaque nature is a liability in domains where transparency is crucial~\citep {10.1145/3442188.3445923, Hong_2020}.


To address the pertinent need for transparency and explainability of their decision-making, the interpretability of machine learning models has emerged as a central focus of recent research~\citep{arrieta2019explainable,SAEED2023110273,räuker2023transparent}.
A popular and well-studied class of interpretability methods is known as \textit{feature attributions}~\citep{ribeiro2016why,lundberg2017unified,sundararajan2017axiomatic}.
Given a model and an input, a feature attribution method assigns scores to input features that reflect their respective importance toward the model's prediction.
A key limitation, however, is that the attribution scores are only as interpretable as the underlying features themselves~\citep{zytek2022need}.

Feature-based explanation methods commonly assume that the given features are already interpretable to the user, but this typically only holds for low-dimensional data.
With high-dimensional data like images and text documents, where the readily available features are individual pixels or tokens, feature attributions are often difficult to interpret~\citep{nauta2023anecdotal}.
The main problem is that features at the individual pixel or token level are often too granular and thus lack clear semantic meaning in relation to the entire input.
Moreover, the important features are also domain-dependent, which means that different attributions are needed for different users.
These factors limit the usefulness of popular feature attribution methods on high-dimensional data.

\input{figures/datasets_summary}

Instead of individual features, people tend to  understand high dimensional data better in terms of semantic collections of low level features, such as regions in an image or phrases in a document. Moreover, for a feature to be useful, it should align with the intuition of \textit{domain experts} in the field.
To this end, an interpretable feature for high-dimensional data should have the following properties.
First, they should encompass a grouping of related low-level features (e.g., pixels, tokens), thus creating high-level features that experts can more easily digest.
Second, these low-level feature groupings should align with domain experts' knowledge of the relevant task, thus creating features with practical relevance. 
We refer to features that satisfy these criteria as \textbf{expert features.}

But how can we obtain such features?
In practice, this process is left to domain experts to identify and provide such features for individual tasks. 
Although experts often have a sense of what the expert features should be, formalizing such features is often non-trivial and difficult.
Moreover, besides formalizing, manually annotating expert features can also be expensive and labor-intensive.
Towards obtaining high-quality features, we ask the following question:
\begin{center}
    Can we automatically measure how well features align with expert knowledge?
\end{center}
To this end, we present the FIX benchmark, a unified evaluation measuring feature interpretability that can capture each individual domain's expert knowledge.
We propose a class of metrics called the \datasetscore and a collection of real-world datasets with expert-designed features.

Our goal is to guide the development of new methods to produce interpretable features by introducing a unified evaluation metric for the expert interpretability of feature groups.
The \dataset datasets (summarized in Figure \ref{fig:datasets_summary}) collectively encompass a diverse array of real-world settings (cosmology, psychology, and medicine) and data modalities (vision, language, and time-series): abdomen surgery safety identification~\citep{madani2022artificial}, chest X-ray classification~\citep{lian2021structure}, mass maps regression~\citep{cosmogrid1}, supernova classification~\citep{lsst}, multilingual politeness classification~\citep{havaldar-etal-2023-comparing}, and emotion classification~\citep{demszky-etal-2020-goemotions, havaldar-etal-2023-multilingual}. 
The challenge lies in unifying all 6 different real-world settings and 3 different data modalities into a \textit{single} framework.
We achieve this with our proposed expert alignment measure \datasetscore, allowing for a benchmark that does not overfit any particular domain.
To our knowledge, while previous work has identified the need for interpretable features ~\citep{zytek2022need, doshivelez2017rigorous}, a benchmark that measures the interpretability of features for real-world experts does not yet exist.
The \dataset benchmark accomplishes this and also serves as a basis for studying, constructing, and extracting expert features.
In summary:
\begin{enumerate}
    \item In collaboration with domain experts, we develop the \textbf{\dataset} benchmark, a collection of 6 curated datasets with metrics for evaluating the explanation inheritability of high-level features.
    Our datasets are taken from real-world settings and covers diverse modalities spanning images, text, and time-series data.
    \footnote{Packaged libraries of code, hugging face data loaders and updates are available at \url{https://brachiolab.github.io/fix/}}
    
    \item We introduce a general feature evaluation metric, \datasetscore, that unifies the different real-world settings of cosmology, psychology, and medicine into a single framework. The criteria for what made features interpretable in each domain were closely informed by real domain experts.

    \item We evaluate commonly used techniques for extracting higher-level features and find that existing methods score poorly on \datasetscore, highlighting the need for developing new general-purpose methods designed to automatically extract expert features.
\end{enumerate}

%% file: figures/datasets_summary.tex
\begin{figure*}[t]
  \centering
  \makebox[\textwidth][c]{\includegraphics[width=\textwidth]{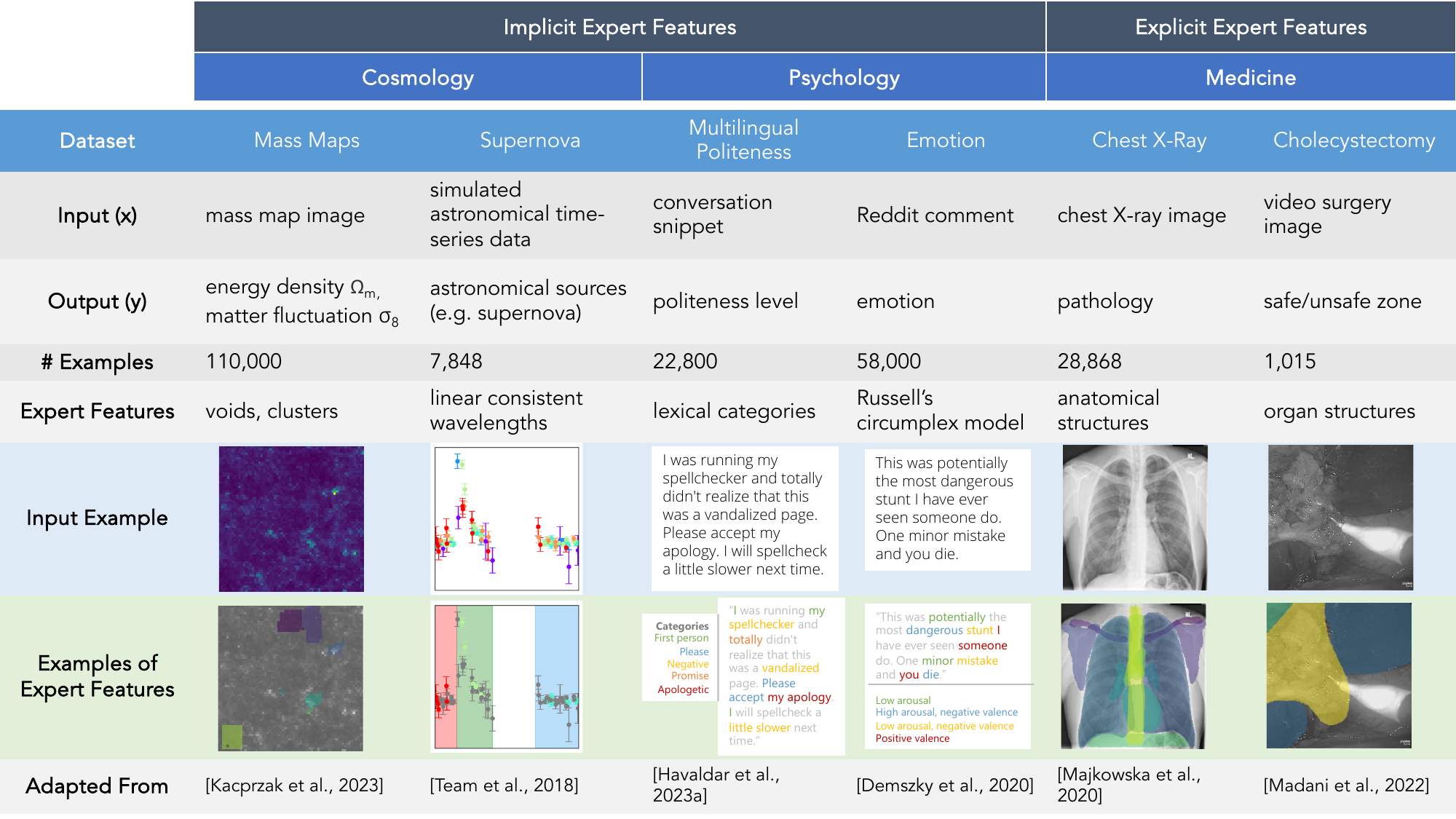}}
  \caption{
    The \dataset benchmark contains 6 datasets across a diverse set of application areas, data modalities, and dataset sizes. For each dataset, we show an example of an input and some example expert features for that input.
    }
  \label{fig:datasets_summary}
\end{figure*}

%% file: sections/related_work.tex
\section{Related Work}
\label{sec:related}

\textbf{Interpretability.}
Interpretability in machine learning is a multifaceted concept that encompasses algorithmic transparency~\citep{shin2019role,rader2018explanations,grimmelikhuijsen2023explaining}, explanation methods~\citep{marcinkevivcs2023interpretable, havaldar2023topex}, and visualization techniques~\citep{choo2018visual,spinner2019explainer,wang2023deepbio}, among other aspects.
In this work, we focus on feature-level interpretability, a central topic in interpretability research~\citep{Hong_2020,nauta2023anecdotal}.
Feature-based methods are popular because they are believed to offer simple, adaptable, and intuitive settings in which to analyze and develop interpretable machine learning workflows~\citep{molnar2020interpretable}.
We refer to~\citep{nauta2023anecdotal,dwivedi2023explainable,weber2023beyond} and the references therein for extensive reviews on feature-based explanations.

\textbf{Application-grounded Evaluation.}
\citet{chaleshtori2023on} extend the work of \citet{doshivelez2017rigorous} to propose a comprehensive taxonomy of evaluating explanations.
Notably, this includes \textit{application-grounded evaluations}, which broadly seek to measure the efficacy of feature-based methods in settings with human users and realistic tasks, such as AI-assisted decision-making.
However, the available literature on application-grounded evaluations is sparse:~\citet{chaleshtori2023on} reviewed over 50 existing NLP datasets and found that only four were suitable for application-grounded evaluations~\citep{deyoung2019eraser,wadden2020fact,koreeda2021contractnli,malik2021ildc}.
A principal objective of the \dataset benchmark is to provide an application-grounded evaluation of feature-based explanations in real-world settings.

\textbf{Feature Generation.}
Because high-quality and interpretable features may not always be available, there is interest in automatically generating them by combining low-level features~\citep{ijcai2017p352,erickson2020autogluontabular,zhang2023openfe}.
Notably, \citet{zhang2023openfe} propose a method for tabular data using the expand-and-reduce framework~\citep{kanter2015deep}.
However, existing generation methods do not necessarily produce interpretable features, and most works focus on tabular data.
The \dataset benchmark aims to address these limitations by providing a setting in which to study and develop methods for interpretable feature generation across diverse problem domains.

\textbf{XAI Benchmarks.} 
There exists a suite of benchmarks for explanations that cover the properties of faithfulness (or fidelity)~\citep{electronics10050593, agarwal2022openxai}, robustness~\citep{DBLP:journals/corr/abs-1806-08049, agarwal2022openxai}, simulatability~\citep{mills2023almanacs}, fairness~\citep{DBLP:journals/corr/abs-2112-04417, agarwal2022openxai}, among others.
Quantus~\citep{hedstrom2023quantus},
XAI-Bench~\citep{DBLP:journals/corr/abs-2106-12543}, OpenXAI~\citep{agarwal2022openxai}, GraphXAI~\citep{agarwal2023evaluating},
and ROAR~\citep{hooker2019benchmark}
are notable open-source implementations that evaluate for such properties.
CLEVR-XAI~\citep{ARRAS202214} and \citet{zhang2023xai} provide benchmarks that combine vision and text.
ERASER~\citep{deyoung2019eraser} is a popular NLP benchmark that unifies diverse NLP datasets of human rationales and decisions.
In general, however, there is a lack of interpretability benchmarks that evaluate feature interpretability in real-world settings --- a gap we aim to address with the \dataset benchmark.

%% file: sections/problem_settings.tex
\section{Expert Feature Extraction}
\label{sec:problem_settings}

\input{figures/IF_extraction}

Feature-based explanation methods require interpretable features to be effective.
For example, surgeons communicate safety in surgery with respect to key anatomical structures and organs, which are interpretable features for surgeons~\citep{strasberg2010rationale,hashimoto2019surgical}.
These interpretable features are a key bridge that can help surgical AI assistants communicate effectively with surgeons.
However, ground-truth annotations for such interpretable features are often expensive and hard to obtain, as they typically require trained experts to manually annotate large amounts of data.
This bottleneck is not unique to surgery, and such challenges motivate us to study the problem of extracting \textit{features interpretable to experts}, or what we call expert features.

Consider a task with inputs from \(\mcal{X} \subseteq \mbb{R}^d\) and outputs in \(\mcal{Y}\).
In the example of surgery, \(\mcal{X}\) may be the set of surgery images, and \(\mcal{Y}\) is the target of where it is safe or unsafe to operate.
We model a higher-level expert feature of input \(x \in \mcal{X}\) as a subset of features represented with a binary mask $g \in \{0,1\}^d$, where $g_i = 1$ if the $i$th feature is included and $g_i = 0$ otherwise.
In surgery, for example, a good high-level feature is one that accurately selects a key anatomical structure or organ from an input \(x\).
The objective of interpretable feature extraction is to find a set of masks \(\hat{G} \subseteq \{0,1\}^d\) that effectively approximates the expert features of \(x\).
That is, each binary mask \(\hat{g} \in \hat{G}\) aims to identify some subset of features meaningful to experts.

However, given a candidate subset of features, how can we judge whether the resulting subset is actually meaningful to experts? To analyze and evaluate potential expert features, we adopt the following \textbf{key guiding principle}: expert features should be \textbf{\emph{designed by experts, for experts}}. Specifically, to ensure broad utility to experts in real world problems, we have designed the FIX benchmark to satisfy the following three properties:  
\begin{enumerate}
    \item \textbf{Formulated by Experts}: Desirable expert features and their corresponding evaluation metrics should be developed by experts and be widely-accepted in their field. In all settings, we work directly with experts to ensure that all of the \dataset datasets and their expert features are well supported and accepted in each domain. 
    \item \textbf{Misalignment of Models and Experts}: 
    We focus the \dataset benchmark on settings where experts by default reason with respect to expert features, but machine learning models typically use low level features. This mismatch is a major communication barrier when explaining model predictions to experts. The FIX settings span problems in medicine, scientific discovery, and social science where experts regularly communicate via expert features, such as organs in surgery, but models are trained in high dimensional inputs, such as high resolution images. 
    \item \textbf{Measure Algorithmic  Progress in Expert  Feature Extraction:}
    The ultimate goal of this benchmark is to guide the develop of novel expert feature extraction methods. to ensure that algorithms are of use to the broader scientific community, solutions should not be overly tailored to any single task. The FIX settings are designed to span a variety of machine learning modalities (vision, language, and time series) and learning problems (clarification, regression, and segmentation). 
\end{enumerate}

In contrast, existing interpretability benchmarks do not closely tie the features to expert knowledge. For example, CLEVR-XAI ~\citep{ARRAS202214}, ERASER ~\citep{deyoung2019eraser}, and ToolQA ~\citep{zhuang2023tool} benchmarks are built synthetically or are typical machine learning benchmarks that do not necessarily align with expert knowledge in practical domains. 
Other benchmarks, such as \citet{ismail2020benchmarking}, DRAC~\citep{qian2024drac}, and FIND~\citep{schwettmann2024find}
are task-specific and do not measure general algorithmic progress across domains. 

\subsection{Measuring Alignment of Extracted Features with Expert Features}

Suppose we are given a function \(\expertalign(\hat{g}, x) \in [0,1]\) that measures how expert-interpretable a group \(\hat{g} \in \{0,1\}^d\) is for input \(x \in \mbb{R}^{d}\).
Such alignment functions for individual groups are common in related tasks, such as in word semantics~\citep{Mathew_2020}, segmentation~\citep{cordts2016cityscapes,abu2018augmented} or object detection~\citep{everingham2010pascal,lin2014microsoft} etc.
The challenge in designing  \(\datasetscore\) is to extend \(\expertalign\) to a \textit{set} of groups \(\hat{G} \subseteq \{0,1\}^d\) while ensuring that individual low-level features are well-covered by \(\hat{G}\).
To do this, we first define how well each low-level feature \(i = 1, \ldots, d\) aligns with respect to \(\hat{G}\) and \(x\) as follows:
\begin{equation}
\label{eqn:featurealign}
    \msc{FeatureAlign}(i, \hat{G}, x)
    = \begin{dcases}
        0, & \text{if \(\hat{G}[i] = \emptyset\)} \\
        \frac{1}{\abs{\hat{G}[i]}}
    \sum_{\hat{g} \in \hat{G}[i]} \expertalign(\hat{g}, x)
        , &\text{otherwise}
    \end{dcases}
\end{equation}
where \(\hat{G}[i] = \{\hat{g} \in \hat{G} : i \in \hat{g}\}\) are the groups of \(\hat{G}\) that cover feature \(i\).
This measures how well, on average, each covering group of \(i\) aligns with the expert criteria of interpretability.
This is to promote that each group of \(\hat{G}[i]\) usefully contributes towards the alignment metric.
We then extend \(\msc{FeatureAlign}\) to all the low-level features to define:
\begin{equation}
    \datasetscore(\hat{G}, x)
    = \frac{1}{d} \sum_{i = 1}^{d} \msc{FeatureAlign}(i, \hat{G}, x)
    \label{eqn:datasetscore}
\end{equation}
where we note that \(\datasetscore\) is parametrized by the particular choice of \(\expertalign\) function.
\(\datasetscore\) can thus be thought of as an average of averages: the expert alignment for each individual feature \(i = 1, \ldots, d\) is averaged over all covers \(\hat{G}[i]\). As a result, this metric has two key strengths regarding feature coverage:
\begin{enumerate}
    \item \textbf{Duplication Invariance at Optimality.} If one extracts perfect expert features (i.e., \(\datasetscore(\hat{G}, x) = 1\) for some \(\hat{G}\) and \(x\)), the \datasetscore cannot be increased further by duplicating expert features. This property ensures that the score cannot be trivially inflated with repeated features. 
    \item \textbf{Encourages Diversity of Expert Features.} Since the score aggregates a value for each feature from $i=1, \dots, d$, adding a new expert feature that does not yet overlap with already extracted features is always beneficial. 
\end{enumerate}

The use of a generic expert alignment function enables the \datasetscore to accommodate a diverse set of applications which fulfills the first desiderata of domain agnostic. To satisfy the third desideratum of expert alignment, \datasetscore includes an expert alignment function customized by experts for each domain. There are two main ways one can specify the $\mrm{\expertalign}$ function: \textit{implicitly} with a score specified by an expert or \textit{explicitly} with annotations from an expert, as shown in Figure~\ref{fig:implicit_vs_explicit}.

\textbf{Case 1: Implicit Expert Alignment.}
Suppose we do not have explicit annotations of expert features for ground truth groups. In this case, we use implicit expert features defined indirectly via a scoring function that measures the quality of an extracted feature. The exact formula of the score is specified by an expert and will depend on the domain and task. Implicit expert features have the advantage of potentially being more scalable than features manually annotated by experts. 

\textbf{Case 2: Explicit Expert Alignment.}
In the case where we do have annotations for expert features $G^{\star}$, we can use a standardized expression for the \datasetscore that measures the best possible intersection with the annotated expert features. 
Then, the expert alignment score of a feature group $\hat{g}$ is
\begin{equation}
\label{eqn:expert_align}
    \expertalign (\hat{g}, x) = \max_{g^{\star} \in G^{\star}(x)} \frac{\abs{\hat{g} \cap g^\star}}{\abs{\hat{g} \cup g^\star}}
\end{equation}
and \(\abs{\cdot}\) counts the number of ones-entries, and \(\cap\) and \(\cup\) are the element-wise conjunction and disjunction of two binary vectors, respectively.
In other words, in the explicit case where the ground-truth expert features are known, alignment amounts to finding the best IoU score among all the expert-defined features \(G^\star\).
Matching intuition, \(\datasetscore\) attains its optimal value at \(\hat{G} = G^\star\):

\begin{theorem}
In the explicit case where \(G^\star\) is known and has full coverage (for all features \(i = 1, \ldots, d\), there exists \(g^\star \in G^\star\) such that \(i \in g^\star\)), we have \(\datasetscore(G^\star, x) = 1\) for all \(x\). \end{theorem}

In this benchmark, the Mass Maps, Supernova, Multilingual Politeness, and Emotion datasets are examples of the implicit features case. On the other hand, the Cholecystectomy and Chest X-ray datasets are examples of the explicit expert features case.

Our goal in FIX is to benchmark general-purpose feature extraction techniques that are \emph{domain agnostic} and do not use the \datasetscore during training. Instead, benchmark challengers can use neural network models trained on the end-to-end tasks to automatically extract features without explicit supervision, which we release as part of the benchmark and discuss further in Appendix~\ref{app:ife}. Annotations for expert features are too expensive to collect at scale for training, while implicit features are by no means comprehensive. The FIX benchmark is intended for evaluation purposes to spur research in general purpose and automated expert feature extraction.



%% file: figures/IF_extraction.tex
\begin{figure}[t]
\begin{center}
\centerline{\includegraphics[width=0.7\columnwidth]{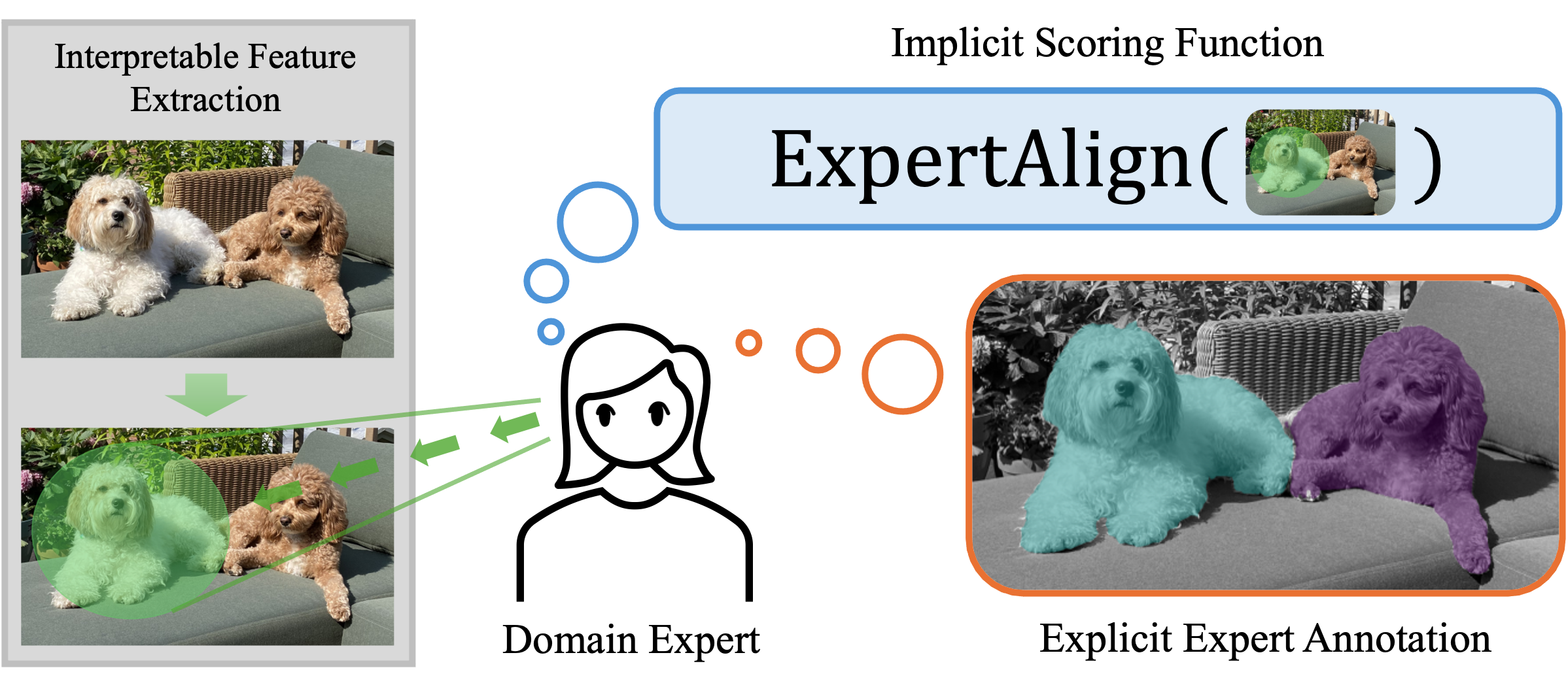}}
\caption{The FIX benchmark allows measuring alignment of extracted features with expert features in different domains, either implicitly with a scoring function or explicitly with expert annotations.}
\label{fig:implicit_vs_explicit}
\end{center}
\end{figure}

%% file: sections/datasets.tex
\section{\dataset Datasets}
\label{sec:datasets}

To develop the FIX benchmark, we curated datasets for expert features designed by experts in accordance with the properties discussed in Section~\ref{sec:problem_settings}. 
In this section, we briefly describe each \dataset dataset in Figure~\ref{fig:datasets_summary}. For each dataset, we provide an overview of the domain task and the problem setup. We then introduce the key expert alignment function that measures the quality of an expert feature, and explain why certain properties incorporated in the expert alignment function are desirable to experts.




%% file: datasets/cosmology.tex
\subsection{Mass Maps Dataset}
\textbf{Motivation.}
A major focus of cosmology is the initial state of the universe, which can be characterized by various cosmological parameters such as \Om{}, which relates to energy density, and \seight{}, which pertains to matter fluctuations~\citep{Abbott_2022}.
These parameters influence what is observable by mass maps, also known as weak lensing maps, which capture the spatial distribution of matter density in the universe.
Although mass maps can be obtained through the precise measurement of galaxies~\citep{y3-massmapping,y3-shapecatalog}, it is not known how to directly measure \Om{} and \seight{}.
This has inspired machine learning efforts to predict the two cosmological parameters from simulations~\citep{ribli2019weak,matilla2020weaklensing,Fluri_2022}.
However, it is hard for cosmologists to gain insights into how to predict \Om{} and \seight{} from black-box ML models.

\textbf{Problem Setup.}
Our dataset contains clean simulations from CosmoGridV1 \citep{cosmogrid1}.
Each input is a one-channel image of size \((66,66)\), where the task is to predict \Om{} and \seight{}.
Here, \Om{} is the average energy density of all matter relative to the total energy density, including radiation and dark energy, while \seight{} describes fluctuations in the distribution of matter \citep{Abbott_2022}.
The dataset has contains  train/validation/test splits of sizes 90,000/10,000/10,000, respectively.

\textbf{Expert Features.}
When inferring \Om{} and \seight{} from the mass maps, we aim to discover which cosmological structures most influence these parameters.
Two types of cosmological structures in mass maps known to cosmologists are voids and clusters \citep{matilla2020weaklensing}.
An example is illustrated in Figure~\ref{fig:massmaps_example}, where voids are large regions that are under-dense relative to the mean density and appear as dark, while clusters are over-dense and appear as bright dots.

To quantify the interpretability of an expert feature in the mass maps, we develop an implicit expert alignment scoring function.
Intuitively, a group that is purely void or purely cluster is more interpretable in cosmology, while a group that is a mixture is less interpretable.
We thus develop the purity metric based on the entropy among void/cluster pixels \citep{zhang2003entropy} weighted by the ratio of interpretable pixels in the expert feature.
We give additional details in Appendix~\ref{app:mass_maps_details}.
\begin{equation}
\expertalign(\hat{g}, x) =
    \mathrm{Purity}_{vc}(\hat{g}, x)\cdot \mathrm{Ratio}_{vc}(\hat{g}, x)
\end{equation}

\input{figures/massmaps_example}

%% file: figures/massmaps_example.tex
\begin{figure}[t]
\centering
\begin{minipage}{0.3\linewidth}
    \centering
    \includegraphics[width=1.0in]{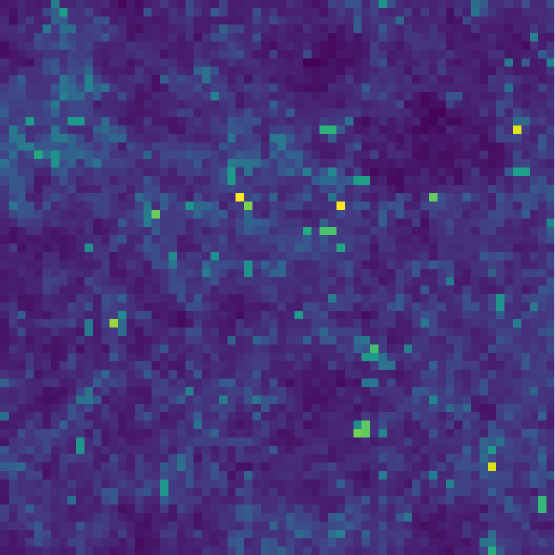}
    \\
    (a) Full map
\end{minipage}\hspace{10pt}
\begin{minipage}{0.3\linewidth}
    \centering
    \includegraphics[width=1.0in]{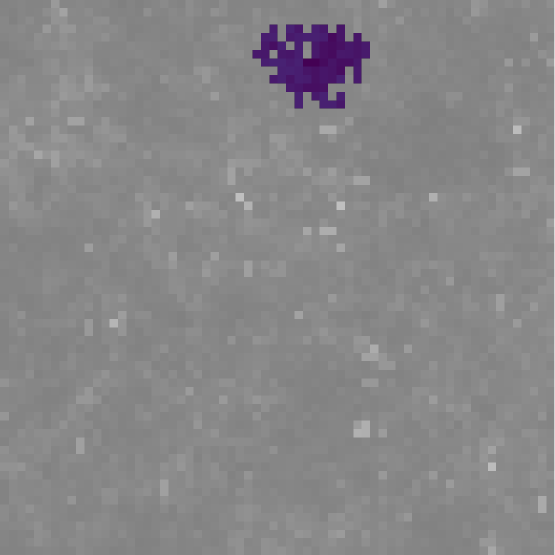}
    \\
    (b) Void
\end{minipage}\hspace{10pt}
\begin{minipage}{0.3\linewidth}
    \centering
    \includegraphics[width=1.0in]{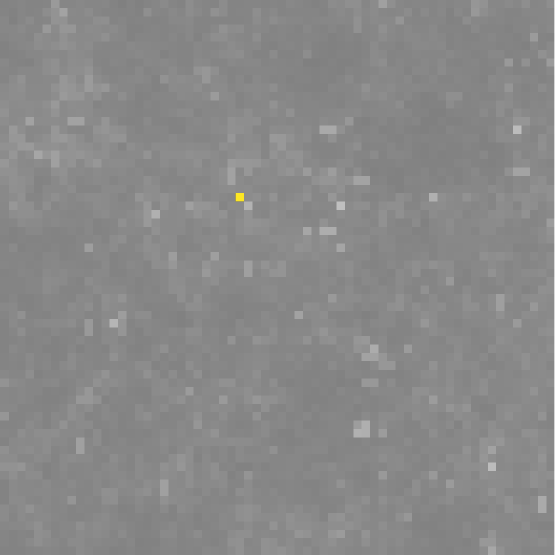}
    \\
    (c) Cluster
\end{minipage}
\caption{
An example with expert features for Mass Maps Regression, showing (a) the full map, (b) a feature with 100\% void, and (c) a feature with 100\% cluster. 
Voids are under-dense large regions that appear to be dark, and clusters are over-dense regions that appear as bright dots.
The purity scores for both void and cluster are 1.
We gray-out the pixels not selected in each feature.
} 
\label{fig:massmaps_example}
\end{figure}

%% file: datasets/supernova.tex
\subsection{Supernova Dataset}
\textbf{Motivation.}
The astronomical time-series classification, as mentioned in~\citep{theplasticcteam2018photometric}, involves categorizing astronomical sources that change over time. 
Astronomical sources include transient phenomena (e.g., supernovae, kilonovae) and variable objects (e.g., active galactic nuclei, Mira variables). 
This task analyzes simulation datasets that emulate future telescope observations from the Legacy Survey of Space and Time (LSST)~\citep{lsst}. 
Given the vastness of the universe, it is essential to identify the time periods that have the most significant impact on the classification of astronomical sources to optimize telescope observations. 
Time periods with no observed data are less useful.
To avoid costly searching over all timestamps for high-influence time periods, we aim to identify significant timestamps that are linearly consistent in specific wavelengths.

\textbf{Problem Setup.}
We take parts of the dataset from the original PLAsTiCC challenge \citep{theplasticcteam2018photometric}.
The input data are simulated LSST observations comprising four columns: observation times (modified Julian days), wavelength (filter), flux values, and flux error. The dataset encompasses 7 distinct wavelengths that work as filters, and the flux values and errors are recorded at specific time intervals for each wavelength. 
The classification task is to predict whether or not each of 14 different astronomical objects exists.
The supernova dataset contains train/validation/test splits of sizes 6274/728/792, respectively.

\textbf{Expert Features.}
A feature with linearly consistent flux for each wavelength is considered more interpretable in astrophysics. An illustration of expert features used for supernova classification is presented in Figure ~\ref{fig:supernova_example}. This example showcases the flux value and error for various wavelengths, each represented by a different color. We colored the timestamp of expert features with the wavelength color with the highest linear consistency score. For timestamps where there are no data points, we do not recognize them as expert features.
We create a linear consistency metric to assess the expert alignment score of a proposed feature in the context of a supernova.
Our linear consistency metric uses $p$, the percentage of data points that display linear consistency, penalized by $d$, the percentage of time stamps containing data points:
\begin{align}
    \mrm{\expertalign} (\hat{g}, x) = \max_{w \in W} p(\hat{g}, x_w)\cdot d(\hat{g}, x_w).
\end{align}
where $W$ is the set of unique wavelength.
Further details are provided in Appendix~\ref{app:supernova_metric}.

\input{figures/supernova_example}

%% file: figures/supernova_example.tex
\begin{figure}[t]
\centering
\begin{minipage}{0.3\linewidth}
    \centering
    \includegraphics[width=1.0\linewidth]{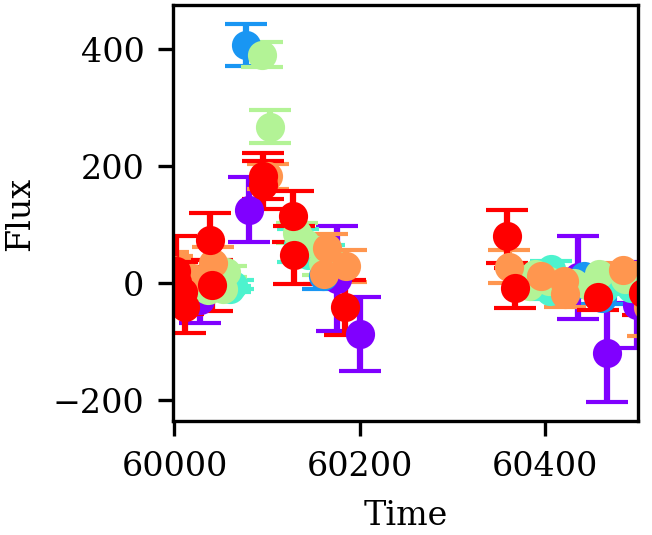}
    \\
\end{minipage}\hspace{10pt}
\begin{minipage}{0.3\linewidth}
    \centering
    \includegraphics[width=1.0\linewidth]{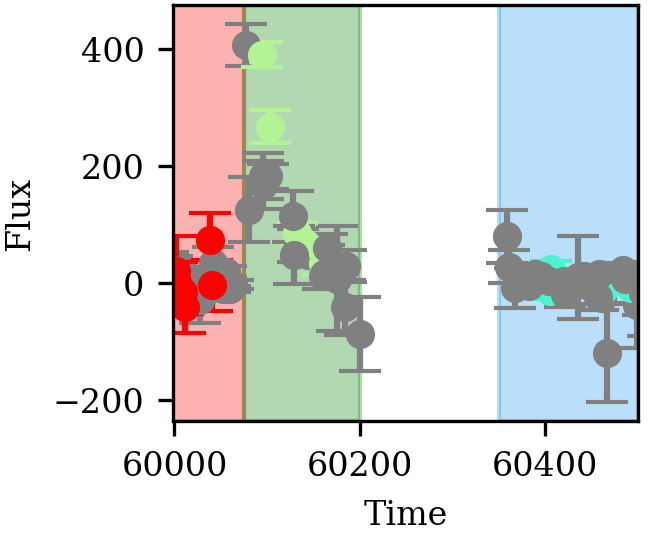}
\end{minipage}
\caption{An example with expert features for supernova classification, showing (left) the original time-series dataset and (right) an example of the interpretable expert feature group.
We highlight the expert feature groups with the highest \expertalign scores.
}

\label{fig:supernova_example}
\end{figure}

%% file: datasets/politeness.tex
\subsection{Multilingual Politeness Dataset} 

\textbf{Motivation.} Different cultures express politeness differently \citep{leech2007politeness, pishghadam2012study}. For instance, politeness in Japan often involves acknowledging the place of others \citep{spencer2016bases}, whereas politeness in Spanish-speaking countries focuses on establishing mutual respect \citep{placencia2017spanish}. Therefore, grounding interpretable features that indicate politeness is \textit{language-dependent}. Previous work from \citet{danescu2013computational} and \citet{mingyang2020politelex} use past politeness research to create lexica that indicate politeness/rudeness in English and Chinese, respectively.
A lexicon is a set of categories where each category contains a curated list of words.
For instance, the English politeness lexicon contains categories like \textit{Gratitude}: ``appreciate'', ``thank you'', et cetera, and \textit{Apologizing}: ``sorry'', ``apologies'', etc.
\citet{havaldar-etal-2023-comparing} expand on these theory-grounded lexica to include Spanish and Japanese.

\textbf{Problem Setup.} The multilingual politeness dataset from \citep{havaldar-etal-2023-comparing} contains 22,800 conversation snippets from Wikipedia's editor talk pages.
The dataset spans English, Spanish, Chinese, and Japanese, and native speakers of these languages have annotated each conversation snippet for politeness level, ranging from -2 (very rude) to 0 (neutral) to 2 (very polite).

\textbf{Expert Features.} When extracting interpretable features for a task like politeness classification across multiple languages, it is useful to ground these features using prior research from communication and psychology. If extracted politeness features from an LLM are interpretable and domain-aligned, they should match what psychologists have determined to be key politeness indicators. 
Examples of expert-aligned features are shown in Table~\ref{tab:emotion_example}.
Concretely, for each lexical category, we use an LLM to embed all the contained words and then average the resulting embeddings to get a set $C$ of $k$ centroids: $C = \{c_1, c_2, \ldots, c_k\}$. See Appendix~\ref{app:politeness_details} for more details. Then, a proposed expert feature $\hat{g}\in\{0, 1\}^d$ indicates whether or not each of the $d$ words $w_1, w_2, ..., w_d \in x$ are included in the feature, and the expert alignment score for the proposed feature $\hat{g}$ can be computed as follows:
\begin{equation}
    \mrm{\expertalign} (\hat{g}, x) = \max_{c \in C} \frac{1}{\abs{\hat{g}}} \sum_{i=1}^d \hat{g}_i \cdot \cos( \text{embedding}(w_i), c)
\end{equation}

%% file: datasets/emotion.tex
\input{figures/emotion_example}

\subsection{Emotion Dataset}

\textbf{Motivation.} Emotion classification involves inferring the emotion (e.g., Joy, Anger, etc.) reflected in a piece of text. 
Researchers study emotion to build systems that can understand emotion and thus adapt accordingly when interacting with human users. 
For extracted features to be useful for such systems, they must be relevant to emotion.
For example, a word like ``puppy'' may be used more frequently in comments labeled with Joy vs. other emotions; therefore, it may be extracted as a relevant feature for the Joy class.
However, this is a spurious correlation --- emotional expression is not necessarily tied to a subject, and comments containing ``puppy'' may also be angry or sad.

\textbf{Problem Setup.} 
The GoEmotions dataset from \citet{demszky-etal-2020-goemotions} contains 58,000 English Reddit comments labeled for 27 emotion categories, or ``neutral'' if no emotion is applicable. 
The input is a text utterance of 1-2 sentences extracted from Reddit comments, and the output is a binary label for each of the 27 emotion categories. 
The dataset contains train/validation/test splits of sizes 43,400/5,430/5,430, respectively.

\textbf{Expert Features.}
Example expert features are shown in Table~\ref{tab:emotion_example}.
To measure how emotion-related a feature is, we use the circumplex model of affect \citep{russell1980circumplex}.
The circumplex model assumes that all emotions can be projected onto the 2D unit circle with respect to two independent dimensions -- 
\textit{arousal} (the magnitude of intensity or activation) and \textit{valence} (how negative or positive).
By projecting features onto the unit circle, we can quantify emotional relations.
In particular, we calculate the following two attributes of the features with a group: (1) their emotional \textit{signal}, i.e., mean distance to the circumplex and (2) their emotional \textit{relatedness}, i.e., mean pairwise distance within the circumplex.
We then calculate the following: $\textrm{Signal}(\hat{g}, x)$, which measures the average Euclidean distance to the circumplex for every projected feature in $\hat{g}$, and $\textrm{Relatedness}(\hat{g}, x)$, which measures the average pairwise distance between every projected feature in $\hat{g}$ (details in Appendix~\ref{app:emotion_details}). For an extracted feature $\hat{g}$, the expert alignment score can then be computed by:
\begin{equation}
    \mrm{\expertalign} (\hat{g}, x) = \mrm{tanh}\parens*{\exp\bracks*{-\mrm{Signal}(\hat{g}, x) \cdot \mrm{Relatedness}(\hat{g}, x)}}
\end{equation}

%% file: figures/emotion_example.tex
\begin{table}[t]
    \centering
    \small
    \begin{tabular}{p{0.4\textwidth}p{0.54\textwidth}}
    \toprule
     \textbf{Example}   &  \textbf{Expert Features with High Alignment } \\
      \midrule
      \multirow{5}{=}{\parbox{0.4\textwidth}{\textit{[Politeness]} I was running my spellchecker and totally didn't realize that this was a vandalized page. Please accept my apology. I will spellcheck a little slower next time.}} & $g_1$ = \textexpertfeatfont{\small{I, my, I}} \\
      & $g_2$ = \textexpertfeatfont{\small{spellchecker, vandalized, little, slower}} \\
      & $g_3$ = \textexpertfeatfont{\small{will}} \\
      & $g_4$ = \textexpertfeatfont{\small{my, apology}} \\
      \\
      \midrule
      \multirow{4}{=}{\parbox{0.4\textwidth}{\textit{[Emotion]} This was potentially the most dangerous stunt I have ever seen someone do. One minor mistake and you die.}} & $g_1$ = \textexpertfeatfont{\small{dangerous, die}} \\
      & $g_2$ = \textexpertfeatfont{\small{potentially, minor}} \\
      & $g_3$ = \textexpertfeatfont{\small{mistake, stunt}} \\
      & $g_4$ = \textexpertfeatfont{\small{I, someone, you}} \\
    \bottomrule \\
    \end{tabular}
    \caption{Examples and expert features with high expert alignment for Multilingual Politeness (top) and Emotion (bottom) settings. These expert features correspond to low distance within the emotion circumplex and high similarity with politeness lexica, respectively.}
    \label{tab:emotion_example}
\end{table}

%% file: datasets/chest_xray.tex
\input{figures/chestx_example}

\subsection{Chest X-Ray Dataset}

\textbf{Motivation.}
Chest X-ray imaging is a common procedure for diagnosing conditions such as atelectasis, cardiomegaly, and effusion, among others.
Although radiologists are skilled at analyzing such images, modern machine learning models are increasingly competitive in diagnostic performance~\citep{ahmad2021reviewing}.
Therefore, ML models may prove useful in assisting radiologists in making diagnoses.
However, in the absence of an explanation, radiologists may only trust the model output if it matches their own predictions.
Moreover, inaccurate AI assistants are shown to negatively affect diagnostic performance~\citep{yu2024heterogeneity}.
To address this problem, explainability could be employed as a safeguard to help radiologists decide whether or not to trust the model.
As such, it is important for machine learning models to provide explanations for their diagnoses.

\textbf{Problem Setup.}
We use the NIH-Google dataset~\citep{majkowska2020chest} available from the TorchXRayVision library~\citep{cohen2022xrv}.
This is a relabeling of the NIH ChestX-ray14 dataset~\citep{wang2017chestx} which improved the quality of the original labels.
It contains 28,868 chest X-ray images labeled for 14 common pathology categories: atelectasis, calcification, cardiomegaly, etc.
We randomly partition the dataset into train/test splits of sizes 23,094/5,774, respectively.
The task is a multi-label classification problem for identifying the presence of each pathology.

\textbf{Expert Features.}
Radiology reports commonly refer to anatomical structures (e.g., spine, lungs), which allows radiologists to perform and communicate accurate diagnoses to patients.
We provide these expert-interpretable features in the form of anatomical structure segmentations.
However, because we could not find datasets with both pathology labels and anatomical segmentations, we used a pre-trained model from TorchXRayVision to generate the structure labelings for each image.
We use explicit expert alignment as described in Equation~\ref{eqn:expert_align} to compute alignment of an extracted feature $\hat{g}$ and the 14 predicted anatomical structure segments, including the left clavicle, heart, etc.
Details of the Chest X-Ray dataset can be found in Appendix~\ref{app:chestxray_details}.

%% file: figures/chestx_example.tex
\begin{figure}[t]

\centering
\begin{minipage}{0.3\linewidth}
    \centering
    \includegraphics[width=1.0in]{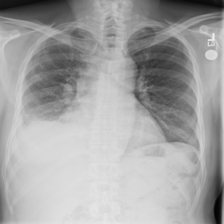}
    \\
    (a) Full image
\end{minipage}\hspace{10pt}
\begin{minipage}{0.3\linewidth}
    \centering
    \includegraphics[width=1.0in]{"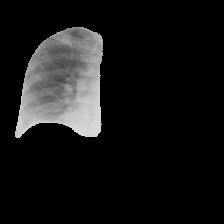"}
    \\
    (b) Right lung
\end{minipage}\hspace{10pt}
\begin{minipage}{0.3\linewidth}
    \centering
    \includegraphics[width=1.0in]{"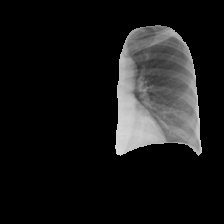"}
    \\
    (c) Left lung
\end{minipage}

\caption{
An example with expert features for Chest X-Ray dataset.
(a) The full X-ray image where the following pathologies are present: effusion, infiltration, and pneumothorax;
(b-c) Expert-interpretable anatomical structures of the left and right lungs.}
\label{fig:chestx_example}
\end{figure}

%% file: datasets/abdomen.tex
\input{figures/cholec_example}

\subsection{Laparoscopic Cholecystectomy Surgery Dataset}

\textbf{Motivation.}
Laparoscopic cholecystectomy (gallbladder removal) is one of the most common elective abdominal surgeries performed in the US, with over 750,000 operations annually~\citep{stinton2012epidemiology}.
A common complication of laparoscopic surgery is bile duct injury, which is associated with an 8-fold increase in mortality~\citep{michael2020safe} and accounts for more than \$1B in US healthcare annual spending~\citep{berci2013laparoscopic}.
Notably, 97\% of such complications result from human visualization errors~\citep{way2003causes}.
The surgery site commonly contains obstructing tissues, inflammation, and other patient-specific artifacts --- all of which may prevent the surgeon from getting a perfect view.
Consequently, there is growing interest in harnessing advanced vision models to help surgeons distinguish safe and risky areas for operation.
However, experienced surgeons rarely trust model outputs due to their opaque nature, while inexperienced surgeons might overly rely on model predictions.
Therefore, any safe and useful machine learning model must be able to provide explanations that align with surgeons' expectations.

\textbf{Problem Setup.}
The task is to identify the safe and unsafe regions for incision.
We use the open-source subset of the data from~\citep{madani2022artificial}, wherein the authors enlist surgeons to annotate surgery video data from the 
M2CAI16 workflow challenge~\citep{stauder2016tum} and Cholec80~\citep{twinanda2016endonet} datasets.
This dataset consists of 1015 annotated images that are randomly split by video sources, with train/test splits of sizes 785/230, respectively.

\textbf{Expert Features.}
In cholecystectomy, it is a common practice for surgeons to identify the \emph{critical view of safety} before performing any irreversible operations~\citep{strasberg2010rationale,hashimoto2019surgical}.
This view identifies the location of vital organs and structures that inform the safe region of operation and is incidentally what surgeons often need as part of an explanation.
We provide these expert-interpretable labels in the form of organ segmentations (liver, gallbladder, hepatocystic triangle).
We use explicit expert alignment as described in Equation~\ref{eqn:expert_align} to compute alignment of an extracted feature $\hat{g}$ and the surgeon-annotated organ labels taken from~\citet{madani2022artificial}.
Details of the Cholecystectomy dataset can be found in Appendix~\ref{app:abdomen_details}.

%% file: figures/cholec_example.tex
\begin{figure}[t]

\centering
\begin{minipage}{0.3\linewidth}
    \centering
    \includegraphics[width=1.0in]{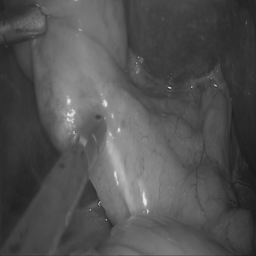}
    \\
    (a) Full image
\end{minipage}\hspace{10pt}
\begin{minipage}{0.3\linewidth}
    \centering
    \includegraphics[width=1.0in]{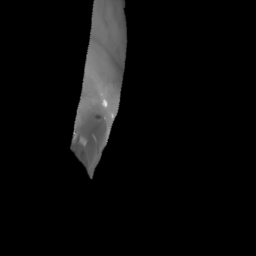}
    \\
    (b) Safe region
\end{minipage}\hspace{10pt}
\begin{minipage}{0.3\linewidth}
    \centering
    \includegraphics[width=1.0in]{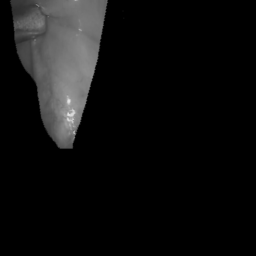}
    \\
    (c) Gallbladder
\end{minipage}

\caption{
An example with expert features of Laparoscopic Cholecystectomy Surgery Dataset:
(a) The view of the surgeon sees;
(b) The safe region for operations;
(c) The gallbladder, a key anatomical structure for the critical view of safety.}
\label{fig:cholec_example}
\end{figure}

%% file: sections/baselines.tex
\section{Baseline Algorithms \& Discussion}
\label{baselines_section}
We evaluate standard techniques widely used within the vision, text, and time series domains to create higher-level features.
We provide a brief summary below, with additional details in Appendix~\ref{app:baselines}.

\input{figures/baseline_scores_simp}

\textbf{Domain-specific Baselines.}
We consider the following domain-centric baselines, which are standard in the literature for the respective domains.
\textit{(Image)}
For image data, we consider three segmentation methods~\citep{kim2024gof}.
Patches~\citep{dosovitskiy2021image} divides the image into grids where each cell is the same size.
Quickshift~\citep{1704833} connects similar neighboring pixels into a common superpixel.
Watershed~\citep{4154796} simulates flooding on a topographic surface.
Segment Anything Model (SAM)~\citep{kirillov2023segment} is a large foundation model for generating image segmentations.
CRAFT~\citep{fel2023craft} generates concept attribution maps.
\textit{(Time-series)}  
For time series data, we take equal size slices of the data across time as patches~\citep{schlegel2021ts}. We use different slice sizes to see how they impact multiple baselines.
We take various slice sizes, such as 5, 10, and 15, separately to evaluate the results of multiple baselines.
\textit{(Text)}
For text data, we present three baselines for extracting features~\citep{rychener2022granularity}.
At the finest granularity, we treat each word as a feature.
The second baseline considers each phrase as a feature. Phrases are comprised of groups of words that are separated by some punctuation in the original text.
At the coarsest granularity, we treat each sentence as a feature.




\textbf{Domain-agnostic Baselines.}
We additionally consider the following domain-agnostic baselines for feature extraction. 
\textit{(Identity)} We combine all elements into one single group.
\textit{(Random)}
We select features at random, up to the maximum baseline results for the group. The group maximum is calculated as: \((\text{group maximum}) \approx (\text{scaling factor}) \times (\text{number of expert features})\).
The size of the distinct expert feature varies depending on the setting, and further details for each setting can be found in Appendix~\ref{app:baselines}. We use a scaling factor of about 1.5 to allow for flexibility.
\textit{(Clustering)}
For images, we first use Quickshift to generate segments and then pass each segment through a feature extractor (ResNet-18 by default).
For time series, we use raw features from each time segment.
We then apply K-means clustering on the extracted/raw features to relabel and merge segments.
For text, we use BERTopic~\citep{grootendorst2022bertopic} to obtain the clusters.
\textit{(Archipelago)}
We adapt the implementation of Archipelago~\citep{tsang2020how} to use ResNet-18 with quickshift for feature extraction.






\textbf{Results and Discussions.}
We show results on the baselines in Table~\ref{tab:baseline_scores}.
For image datasets, Quickshift has the best performance compared to Patch and Watershed on both the Cholecystectomy dataset and the Chest X-ray dataset, since they have natural images.
All baselines perform similarly for the Mass Maps dataset.
That the range of mass maps is different from other tasks is potentially because they are not natural images, but rather similar to topographic surfaces, and also the implicit ground truth expert features do not have full coverage.
For the Supernova time-series dataset, larger slices score yield higher expert alignment scores.
For both Multilingual Politeness and Emotion datasets, individual words appear to be the most expert-aligned features. Generally, however, we see that the domain-agnostic neural baselines tend to also perform better than or close to the best domain-centric baseline. The main benefit of using a neural approach is that it can more easily automatically discover relevant features.

%% file: figures/baseline_scores_simp.tex
\begin{table}[t]

\centering
\setlength{\tabcolsep}{4pt}
\scriptsize
\begin{tabular}{p{1.1cm}|p{1.2cm}ccc|p{1cm}c|p{1cm}cc}
\toprule
& \multicolumn{4}{c|}{Vision} & \multicolumn{2}{c|}{Time Series} & \multicolumn{3}{c}{Language} \\ 
   & \textbf{Method} 
    & \textbf{Cholec}
    & \textbf{ChestX}
    & \textbf{MassMaps} & \textbf{Method} & \textbf{Supernova} & \textbf{Method} & \textbf{Politeness} & \textbf{Emotion}\\
    \cmidrule(lr){1-10}
    \multirow{7}{1cm}{\textit{Domain-specific}} &Identity
       & 0.4648 
       & 0.2154 
       & 0.5483 
       &Identity & 0.0152
       & Identity
       & 0.6070 
       & 0.0103 
       \\
    &Random
        & 0.1084 
        & 0.0427 
        & 0.5505 
        &Random & 0.0358 
        & Random
        & 0.6478
        & 0.0303
        \\
    &Patch 
        & 0.0327 
        & 0.0999 
        & 0.5555
        &Slice 5
        & 0.0337
        & Words
        & 0.6851 
        & 0.1182
        \\
    &Quickshift
        & 0.2664 
        & 0.3419 
        & 0.5492  
        &Slice 10
        & 0.0555
        & Phrases
        & 0.6351 
        & 0.0198 
        \\
    &Watershed
        & 0.2806 
        & 0.1452 
        & 0.5590 
        &Slice 15
        & 0.0554
        & Sentences
        & 0.6109 
        & 0.0120 
        \\
    &SAM
        & 0.3642 
        & 0.3151 
        & 0.5521 & & & & &
        \\
    &CRAFT
        & 0.0278 
        & 0.1175 
        & 0.3996 & & & & &
        \\
    \cmidrule(lr){1-10}
    \multirow{2}{1cm}{\textit{Domain-agnostic}} &Clustering
        & 0.2839 
        & 0.2627 
        & 0.5515 
        &Clustering
         & 0.2622 
         & Clustering 
        & 0.6680 
        & 0.0912 
        \\
    &Archipelago
        & 0.3271 
        & 0.2148 
        & 0.5542 
        &Archipelago
        & 0.2574 
        & Archipelago
        & 0.6773  
        & 0.0527 
        \\
\bottomrule
\end{tabular}
\caption{Baselines scores of different \dataset settings.
We report the mean score and give a more comprehensive table in Appendix~\ref{app:baselines_details}. 
We describe baseline implementations in Section~\ref{baselines_section}.
One thing to note is that \datasetscore is not comparable for different tasks (e.g. between Mass Maps and Supernova) as the data and specific expert alignment metrics are different for different tasks.
}
\label{tab:baseline_scores}
\end{table}

%% file: sections/conclusion.tex
\section{Conclusion}
\label{sec:conclusion}
We propose \dataset, a curated benchmark of datasets with evaluation metrics for extracting expert features in diverse real-world settings.
Our benchmark addresses a gap in the literature by providing researchers with an environment to study and automatically extract interpretable features for experts, designed by experts. 


\textbf{Limitations and Future Work.}
The FIX benchmark is not an exhaustive specification of all expert features, and may fail to capture others types. The ones we included are generally non-controversial and well-accepted by the domain’s expert community, but we can foresee that there are cases where this may not be true. Dealing with potential conflicting expert opinions may need a more nuanced approach, which is left for future work to address. 
Furthermore, although we cover cosmology, psychology, and medicine domains in this work, the metrics for these domains may not be appropriate for all settings. We encourage prospective users to consider and implement metrics most appropriate to their particular settings.
Future work includes the development of new, general purpose techniques that can extract expert features from data and models without supervision. 
Additionally, future work could also include training machine learning models on just the features that are deemed to be aligned with domain experts and reporting the accuracy of the trained models.

%% file: sections/broader_impact.tex
\section*{Broader Impact and Ethics Statement}
\label{sec:impact}
The goal of the FIX benchmark is to enable researchers and practitioners to develop more transparent machine learning systems that are applicable in real-world problems.
However, because our datasets contain text scraped from Internet forums, as well as visuals of human anatomy, it is possible that some contents may be considered objectionable.
It is possible that such objectionable content may be misused, but we do not believe that our datasets would be of particular interest to malicious users because dedicated natural-language toxicity and more graphic medical datasets exist.

%% file: sections/acknowledgement.tex
\section*{Acknowledgment}
\label{sec:acknowledgement}
This research was partially supported by a gift from AWS AI to Penn Engineering’s ASSET Center for Trustworthy AI, by ASSET Center Seed Grant, ARPA-H program on
Safe and Explainable AI under the award D24AC00253-00, by NSF award CCF 2442421, and by funding from the Defense Advanced Research Projects Agency’s (DARPA)
SciFy program (Agreement No. HR00112520300). The views expressed are those of the author and do not reflect the official policy or position of the Department of Defense
or the U.S. Government.

%% file: sections/appendix.tex
\newpage

\section{Dataset Details}
\label{dataset_links}
All datasets and their respective Croissant metadata records and licenses are available on HuggingFace at the following links.
\begin{itemize}[leftmargin=16pt,topsep=1pt,noitemsep]
    
        
    
    
        
    
    
\item \textbf{Mass Maps}:
    
        \quad {\small \url{\massmapsurl}}
        
    \item \textbf{Supernova}:
    
        \quad {\small \url{\supernovaurl}}
    \item \textbf{Multilingual Politeness}:
    
        \quad {\small \url{\politenessurl}}
    \item \textbf{Emotion}:
        
        \quad {\small \url{\emotionurl}}
    \item \textbf{Chest X-Ray}:
    
        \quad {\small \url{\chestxurl}}

    \item \textbf{Laparoscopic Cholecystectomy Surgery}:
    
        \quad {\small \url{\cholecurl}}
\end{itemize}

\label{app:dataset_details}

\input{appendix/dataset_details/cosmology}

\input{appendix/dataset_details/supernova}

\input{appendix/dataset_details/politeness}

\input{appendix/dataset_details/emotion}

\input{appendix/dataset_details/chest_xray}

\input{appendix/dataset_details/abdomen}

\section{Interpretable Feature Extraction Details}
\label{app:ife}
\input{appendix/ife}

\section{Baselines Details}
\label{app:baselines_details}

The \dataset benchmark is publicly available at: \url{https://brachiolab.github.io/fix/}
\label{app:baselines}
\input{appendix/baselines}

\section{Representative Examples of Extracted Features.}
\label{app:dataset_example_features}
Here, we include representative examples of features extracted by existing baseline methods, along with commentary on how they differ from expert-aligned features.

\input{appendix/dataset_example_features/cosmology}

\input{appendix/dataset_example_features/supernova}

\input{appendix/dataset_example_features/politeness}

\input{appendix/dataset_example_features/emotion}

\input{appendix/dataset_example_features/chest_xray}

\input{appendix/dataset_example_features/abdomen}

\section{Adding a New Setting.}
\label{app:add_setting}
\input{appendix/add_setting}

\section{Compute Resources}
\label{app:compute}
\input{appendix/compute}

\section{Safeguards}
\label{app:safeguards}
\input{appendix/safeguards}

\section{Datasheets}
\input{appendix/datasheet}

\section{Author Statement}
We bear all responsibility for any potential violation of rights, etc., and confirmation of data licenses.

%% file: appendix/dataset_details/cosmology.tex
\subsection{Mass Maps Dataset}
\label{app:mass_maps_details}
\paragraph{Problem Setup.}
We randomly split the data to consist of 90,000 train and 10,000 validation maps and maintain the original 10,000 test maps.
We follow the post-processing procedure in \citet{y3-massmapping,you2023sumofparts} for low-noise maps.
Following previous works \citep{ribli2019weak,matilla2020weaklensing,Fluri_2022,you2023sumofparts}, we use a CNN-based model for predicting \Om{} and \seight{}.

\paragraph{Metric.}
Let $x\in\mathbb{R}^d$ be the input mass map with $d=H\times W$ pixels, and $g\in\{0, 1\}^{d}$ be a boolean mask $g$ that describes which pixels belong to the group, where $g_i=1$ if the $i$th pixel belongs to the group, and 0 otherwise. 

We can compute the purity score of each group to void and cluster.
We say a pixel is a void (underdensed) pixel if its intensity is below $0$, and a cluster (overdensed) pixel if its intensity is above $3\sigma(x)$, following previous works \citep{matilla2020weaklensing,you2023sumofparts}.
We first compute the proportion of void pixels and cluster pixels in feature $g$
\begin{gather}
    P_{v}(g, x) = \frac{\sum_{i=1}^d \mathbbm{1}[g_i x_i < 0]}{g^\intercal \mathbf{1}}, \quad\quad P_{c}(g, x) = \frac{\sum_{i=1}^d \mathbbm{1}[g_i x_i > 3\sigma (x)]}{g^\intercal \mathbf{1}}
\end{gather}
where $\mathbf{1}\in 1^d$ is the identity matrix, the numerators count the number of underdensed or overdensed pixels, and $g^\intercal \mathbf{1}$ is the number of pixels in the feature.
In practice, we add a small $\epsilon=10^{-6}$ to $P_v$ and $P_c$ and renormalize them, to avoid taking the log of 0 later.
Next, we compute the proportion of pixels that are void or cluster, only among the void/cluster pixels:
\begin{gather}
    P_{v}'(g, x) = \frac{P_v(g, x)}{P_v(g, x)+P_c(g, x)}, \quad\quad P_{c}'(g, x) = \frac{P_c(g, x)}{P_v(g, x)+P_c(g, x)}
\end{gather}
Then, we compute the $\expertalign$ score for the predicted feature $\hat{g}$ by computing the void/cluster-only entropy reversed and scaled to $[0,1]$, weighted by the percentage of void/cluster pixels among all pixels.
\begin{equation}
    \begin{aligned}
    \mathrm{Purity}_{vc}(\hat{g}, x) = &\frac{1}{2}(2 + P_{v}'(\hat{g}, x) \log_2 P_{v}'(\hat{g}, x) + P_{c}'(\hat{g}, x) \log_2 P_{c}'(\hat{g}, x))
\end{aligned}
\end{equation}
where $-(P_{v}'(\hat{g}, x) \log_2 P_{v}'(\hat{g}, x) + P_{c}'(\hat{g}, x) \log_2 P_{c}'(\hat{g}, x))$ is the entropy computed only on void and cluster pixels, a close to 0 score indicating that the interpretable portion of the feature is mostly void or cluster.
$\mathrm{Purity}_{vc}(\hat{g}, x)$ is 0 if among the pixels in the proposed feature that are either void or cluster pixels, half are void and half are cluster pixels, and 1 if all are void or all are cluster pixels, regardless of how many other pixels there are in the proposed feature.

We also have the ratio
\begin{equation}
    \begin{aligned}
        \mrm{Ratio}_{vc}(\hat{g}, x) = (P_v(\hat{g}, x) + P_c(\hat{g}, x))
    \end{aligned}
\end{equation}
which is the total proportion of the feature that is any interpretable feature type at all.

We then have our $\expertalign$ for Mass Maps:
\begin{equation}
    \expertalign(\hat{g}, x) =
    \mathrm{Purity}(\hat{g}, x) \cdot \mathrm{Ratio}(\hat{g}, x) 
\end{equation}
which is then 0 when all the pixels in the feature are neither void or cluster, and 1 if all pixels are void pixels or all pixels are cluster pixels, and somewhere in the middle if most pixels are void or cluster pixels but there is a mix between both.

%% file: appendix/dataset_details/supernova.tex
\subsection{Supernova Dataset}
\label{app:supernova_details}
\textbf{Problem Setup.}
We extracted data from the PLAsTiCC Astronomical Classification challenge \citep{theplasticcteam2018photometric}. \footnote{\url{https://www.kaggle.com/c/PLAsTiCC-2018}} PLAsTiCC dataset was designed to replicate a selection of observed objects with type information typically used to train a machine learning classifier. The challenge aims to categorize a realistic simulation of all LSST observations that are dimmer and more distorted than those in the training set. The dataset contains 15 classes, with 14 of them present in the training sample. The remaining class is intended to encompass intriguing objects that are theorized to exist but have not yet been observed. 

In our dataset, we split the original training set into 90/10 training/validation, and the original test set was uploaded unchanged. We made these sets balanced for each class. The class includes objects such as tidal disruption event (TDE), peculiar type Ia supernova (SNIax), type Ibc supernova (SNIbc), and kilonova (KN).
The dataset contains four columns: observation times (modified Julian days, MJD), wavelength (filter), flux values, and flux error. Spectroscopy measures the flux with respect to wavelength, similar to using a prism to split light into different colors.

Due to the expected high volume of data from upcoming sky surveys, it is not possible to obtain spectroscopic observations for every object. However, these observations are crucial for us. Therefore, we use an approach to capture images of objects through different filters, where each filter selects light within a specific broad wavelength range. The supernova dataset includes 7 different wavelengths that are used. The flux values and errors are recorded at specific time intervals for each wavelength. These values are utilized to predict the class that this data should be classified into.
\label{app:supernova_metric}
\paragraph{Metric.}
We use the following expert alignment metric to measure if a group of features is interpretable:
\begin{align}
    \mrm{\expertalign} (\hat{g}, x) = \max_{w \in W} \mrm{Linear Consistency} (\hat{g}, x_w)
\end{align}
where $W$ is the set of unique wavelength, $\hat{g}$ is the feature group, and $x_w$ is the subset of $x$ within wavelength $w$. 
In the supernova setting, there are three parameters: $\epsilon$, the parameter for how much standard deviation $\sigma$ is allowed, window size $\lambda$ and the step size $\tau$. 
Therefore, we formulate the $\mrm{Linear Consistency}$ function as follows:
\begin{align}
    \mrm{\centering Linear Consistency} (\hat{g}, x_w) = p(\hat{g}, x_w)\cdot d(\hat{g}, x_w)
\end{align}
$p(\hat{g}, x_w)$ is the percentage of data points that display linear consistency, penalized by $d(\hat{g}, x_w)$, which is the percentage of time steps containing data points.

Let $\beta(x, y) = \argmin_{\beta} (X^T  \beta - y)^2$, where $X = \begin{bmatrix} x & 1 \end{bmatrix}$ and $\beta = \begin{bmatrix} \beta_1 & \beta_0 \end{bmatrix}$. Here, $\beta_1$ is the slope and $\beta_0$ is the intercept.  $M$ is the number of data points in $x_w$, and $\hat{y}_{w, i} = {x}_{w, i} \cdot \beta$. Then, we have
\begin{align}
    &\mrm p(\hat{g}, x_w) =  \frac{1}{M} \sum_{i=1}^M \mathbbm{1}[\hat{y}_{w, i} \in [{y}_{w, i} - \epsilon \cdot \omega_{w, i}, {y}_{w, i} + \epsilon \cdot \omega_{w, i}] ]
\end{align}

Let $t_1, ..., t_N$ be time steps at step size intervals. Then $t_i = t_{start} + i * \tau$, and $N$ is the number of time steps. We also have
\begin{align}
    &\mrm d(\hat{g}, x_w) =  \frac{1}{N} \sum_{i=1}^N \mathbbm{1}[\exists_i : {x}_{w, i} \in [{t}_{i} , {t}_{i} + \lambda] ]
\end{align}

A higher $\mrm{\expertalign} (\hat{g}, x)\in[0,1]$ value means the flux slope at each wavelength is consistently linear and there are not many time intervals without data.

%% file: appendix/dataset_details/politeness.tex
\subsection{Multilingual Politeness Dataset}
\label{app:politeness_details}
\paragraph{Problem Setup.} 
This politeness dataset from \citet{havaldar-etal-2023-multilingual} is intended for politeness classification, and would likely be solved via a fine-tuned multilingual LLM. Namely, this would be a regression task, using a trained LLM to output the politeness level of a given conversation snippet as a real number ranging from -2 to 2.

The dataset is accompanied by a theory-grounded politeness lexica. Such lexica built with domain expert input have been promising for explaining style~\citep{danescu2013computational}, culture~\citep{havaldar-etal-2024-building}, and other such complex multilingual constructs.

\paragraph{Metric.}
Assume a theory-grounded Lexica $L$ with $k$ categories: $L = {\ell_1, \ell_2, ... \ell_k}$, where each set $\ell_i \subseteq \mathcal{W}$, where $\mathcal{W}$ is the set of all words. For each category, we use an LLM to embed all the contained words and then average the resulting embeddings, to get a set $C$ of $k$ centroids: $C = {c_1, c_2, ... c_k}$. We define this formally as:
\begin{equation}
    C: \left \{ \frac{1}{|\ell_i|} \sum_{w \in l_i} \text{embedding}(w) \text{ for all } i \in [1, k]\right \}
\end{equation}

For a group $\hat{g}$ containing words $w_1, w_2, ...$, the group-level expert alignment score can be computed as follows:
\begin{equation}
    \mrm{\expertalign} (\hat{g}, x) = \max_{c \in C} \frac{1}{|\hat{g}|}\sum_{w \in \hat{g}} \cos( \text{embedding}(w), c)
\end{equation}

Note that each language has a different theory-grounded lexicon, so we calculate a unique domain alignment score for each language.

%% file: appendix/dataset_details/emotion.tex
\subsection{Emotion Dataset}
\label{app:emotion_details}

\paragraph{Problem Setup.}
This dataset is intended for emotion classification and is currently solved with a fine-tuned LLM \citep{demszky-etal-2020-goemotions}. Namely, this is a classification task where an LLM is trained to select some subset of 28 emotions (including neutrality) given a 1-2 sentence Reddit comment. 

\paragraph{Projection onto the Circumplex.}

\input{figures/axis_definitions}

To define the valence and arousal axes, we first generate four axis-defining points by averaging the contextualized embeddings ("I feel [emotion]") of the emotions listed in Table~\ref{tab:axis_definitions}. This gives us four vectors in embedding space -- positive valence ($\Vec{v}_{\mrm{pos}}$), negative valence($\Vec{v}_{\mrm{neg}}$), high arousal($\Vec{a}_{\mrm{high}}$), and low arousal($\Vec{a}_{\mrm{low}}$). We mathematically describe our projection function below:
\begin{enumerate}
\vspace{-0.1cm}
    \itemsep=0em
    \item We define the valence axis, $V$, as $\Vec{v}_{\mrm{pos}} - \Vec{v}_{\mrm{neg}}$ and the arousal axis, $A$, as $\Vec{a}_{\mrm{high}} - \Vec{a}_{\mrm{low}}$. We then normalize $V$ and $A$ and calculate the origin as the midpoints of these axes: $(\Vec{v}_{\mrm{middle}}, \Vec{a}_{\mrm{middle}})$. 
    \item We then scale the axes so $\Vec{v}_{\mrm{pos}}$, $\Vec{v}_{\mrm{neg}}$, $\Vec{a}_{\mrm{high}}$, and $\Vec{a}_{\mrm{low}}$ anchor to $(1,0)$, $(-1,0)$, $(0,1)$, and $(0,-1)$ respectively. This enforces the circumplex to be a unit circle in the valence-arousal plane.
    \item We compute the angle $\theta$ between the valence-arousal axes by solving $\cos\theta =  \frac{{V}\cdot{A}}{\norm{V}\cdot\norm{A}}$
    \item For each embedding vector $\Vec{x}$ in the set $\{x_i\}_{i=1}^n$ we want to project into our defined plane, we compute the valence and arousal components for $x_i$ as follows: \\
    $x_i^{v}= (x_i - \Vec{v}_{\mrm{middle}})\cdot \Vec{V}$\\
    $x_i^{a}= (x_i - \Vec{a}_{\mrm{middle}})\cdot \Vec{A}$.
    \item We calculate the x and y coordinates to plot, enforcing orthogonality between the axes: \\
    $\Tilde{x_i^{v}} = x_i^{v} - x_i^{a}\cdot \cos\theta$\\
    $\Tilde{x_i^{a}} = x_i^{a} - x_i^{v}\cdot \cos\theta$
    \item Finally, we plot $(\Tilde{x_i^{v}}, \Tilde{x_i^{v}})$ in the Valence-Arousal plane. We then calculate the shortest distance from $(\Tilde{x_i^{v}}, \Tilde{x_i^{v}})$ to the circumplex unit circle.
\end{enumerate}

\begin{figure}[ht]
\begin{center}
\centerline{\includegraphics[width=0.6\columnwidth]{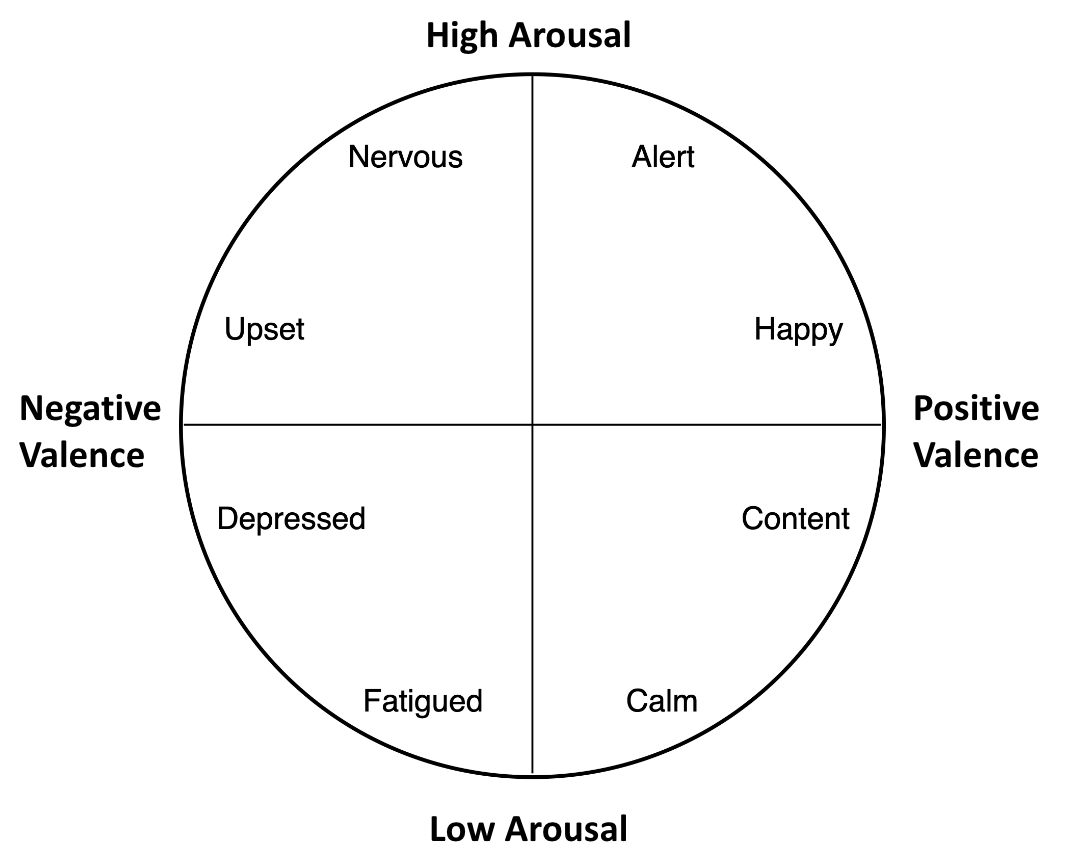}}
\caption{The circumplex model of affect \cite{russell1980circumplex}.}
\label{fig:circumplex}
\end{center}
\vskip -0.2in
\end{figure}

\paragraph{Metric.}

We calculate the following two values for a proposed feature $\hat{g}$ containing words $w_1, w_2, ...$, where $n$ is the number of words in $\hat{g}$:
\begin{align}
    \mrm{Signal}(\hat{g})
        &= \frac{1}{n}\sum_{w \in \hat{g}} \left | \left \|\mrm{Proj}(w) \right \|_2 - 1 \right | \\
    \mrm{Relatedness}(\hat{g})
        &= \frac{1}{n^2}\sum_{i}^{n} \sum_{j}^{n} \|\mrm{Proj}(w_i) - \mrm{Proj}(w_j)\|_2
\end{align}
where $\mrm{Signal}(\hat{g}, x)$ measures the average Euclidean distance to the circumplex for every projected feature in $\hat{g}$, and $\mrm{Relatedness}(\hat{g}, x)$ measures the average pairwise distance between every projected feature in $\hat{g}$. We formalize the expert alignment metric as follows. 
For a group $\hat{g}$, the expert alignment score can be computed by:
\begin{equation}
    \mrm{\expertalign} (\hat{g}, x) = \mrm{tanh}\parens*{\exp\bracks*{-\mrm{Signal}(\hat{g}, x) \cdot \mrm{Relatedness}(\hat{g}, x)}}
\end{equation}

%% file: figures/axis_definitions.tex
\begin{table}[h]
\centering
\small
\begin{tabular}{ll}
    \toprule
    \textbf{Axis Anchor} & \textbf{Russell Emotions} \\ \midrule
    Positive valence (PV) &  \makecell[l]{Happy, Pleased, Delighted, Excited, Satisfied}  \\ 
    Negative valence (NV) &  \makecell[l]{Miserable, Frustrated, Sad, Depressed, Afraid}  \\
    High arousal (HA) &  \makecell[l]{Astonished, Alarmed, Angry, Afraid, Excited}  \\
    Low arousal (LA) &  \makecell[l]{Tired, Sleepy, Calm, Satisfied, Depressed}  \\
\bottomrule
\end{tabular}
\vspace{0.25cm}
\caption{Emotions used to define the valence and arousal axis anchors for projection into the Valence-Arousal plane. We select the 5 emotions from the circumplex closest to each axis point.}
\label{tab:axis_definitions}
\end{table}

%% file: appendix/dataset_details/chest_xray.tex
\subsection{Chest X-Ray Dataset}
\label{app:chestxray_details}

We used datasets and pretrained models from TorchXRayVision~\citep{cohen2022xrv}.\footnote{\url{https://github.com/mlmed/torchxrayvision}}
In particular, we use the NIH-Google dataset~\citep{majkowska2020chest}, which is a relabeling of the NIH ChestX-ray14 dataset~\citep{wang2017chestx}.
This dataset contains 28,868 chest X-ray images labeled for 14 common pathology categories, with a train/test split of 23,094 and 5,774.
We additionally used a pre-trained structure segmentation model to produce 14 segmentations.
The task is a multi-label classification problem for identifying the presence of each pathology.
The 14 pathologies are:
\begin{quote}
    Atelectasis,
    Cardiomegaly,
    Consolidation,
    Edema,
    Effusion,
    Emphysema,
    Fibrosis,
    Hernia,
    Infiltration,
    Mass,
    Nodule,
    Pleural Thickening,
    Pneumonia,
    Pneumothorax
\end{quote}

The 14 anatomical structures are:
\begin{quote}
    Left Clavicle,
    Right Clavicle,
    Left Scapula,
    Right Scapula,
    Left Lung,
    Right Lung,
    Left Hilus Pulmonis,
    Right Hilus Pulmonis,
    Heart,
    Aorta,
    Facies Diaphragmatica,
    Mediastinum,
    Weasand,
    Spine
\end{quote}

%% file: appendix/dataset_details/abdomen.tex
\subsection{Laparoscopic Cholecystectomy Surgery Dataset}
\label{app:abdomen_details}

We use the open-source subset of the data from~\citep{madani2022artificial}, which consists of surgeon-annotated video data taken from the 
M2CAI16 workflow challenge~\citep{stauder2016tum} and Cholec80~\citep{twinanda2016endonet} datasets.
The task is to identify the safe/unsafe regions of where to operate.
Specifically, each pixel of the image has one of three labels: background, safe, or unsafe.
The expert labels provide each pixel with one of four labels: background, liver, gallbladder, and hepatocystic triangle.

%% file: appendix/ife.tex
\input{figures/fix_pipeline}

%% file: figures/fix_pipeline.tex
\begin{figure}[ht!]
\begin{center}
\centerline{\includegraphics[width=0.7\columnwidth]{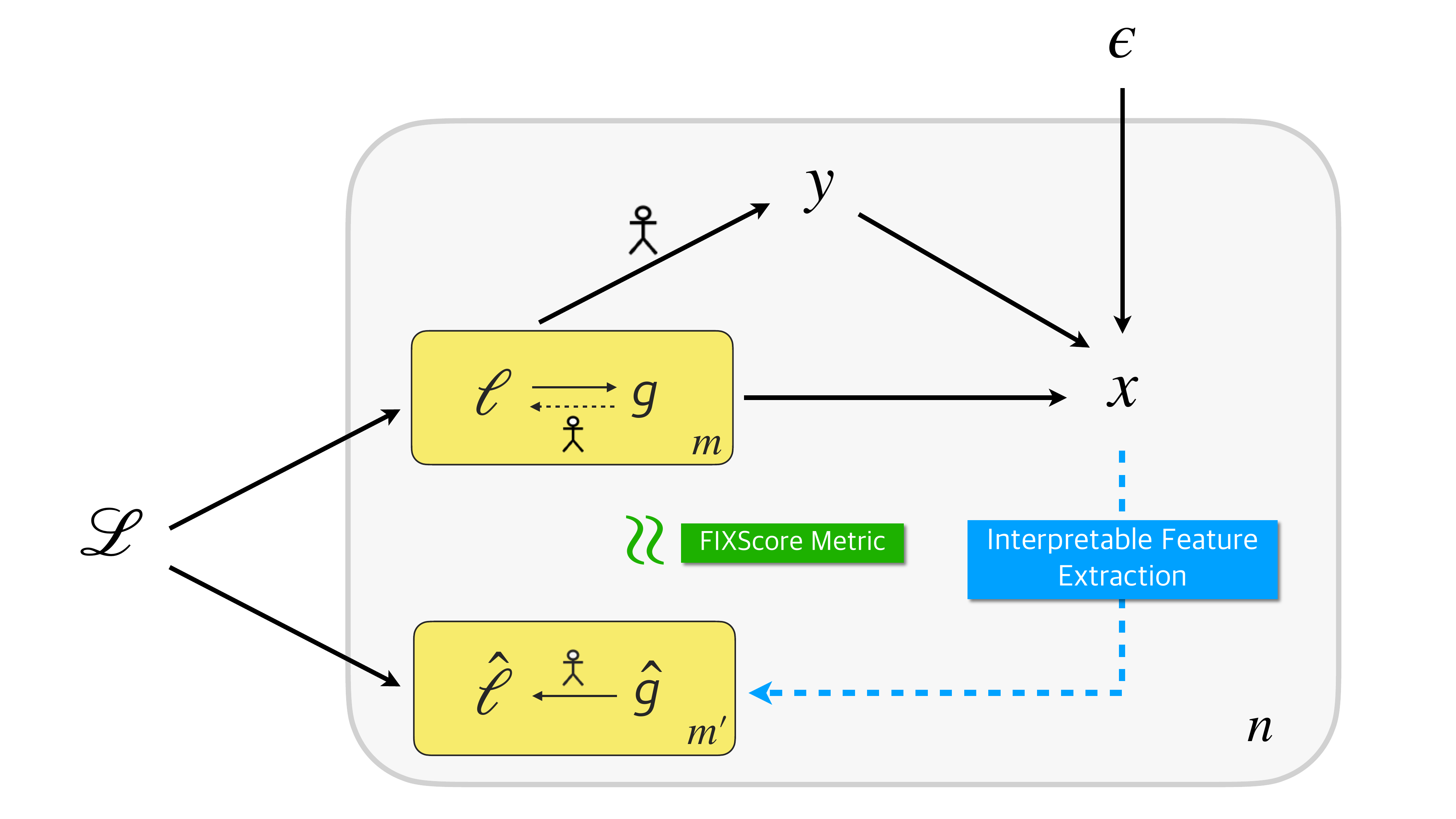}}
\caption{We illustrate a graphical model representing the Interpretable Feature Extraction pipeline for a given \dataset dataset, with \datasetscore metric in its general form. There are $m$ true feature groups $g$ and $m$ latent features $\ell$, and $m'$ proposed feature groups $\hat{g}$ and $m'$ proposed latent features $\hat{\ell}$. $m$ does not have to equal $m'$. Moreover, $n$ indicates the number of examples in the dataset. The person figure on near the closest arrow indicates that a domain expert would be able to infer the variable on the right-hand side of the arrow from the variable on the left-hand side arrow. In addition, $\epsilon$ is included to account for noise. }
\label{fig:fix_pipeline}
\end{center}
\vskip -0.2in
\end{figure}

Figure \ref{fig:fix_pipeline} illustrates a graphical model representing the Interpretable Feature Extraction pipeline for a given \dataset dataset.

%% file: appendix/baselines.tex
\paragraph{Bootstrapping.}
For each setting's baselines experiments, we use a bootstrapping method (with replacement) to estimate the standard deviation of the sample means of \datasetscore.

\paragraph{Group Maximum.}

For the number of groups, we take the scaling factor multiplied by the size of the distinct expert feature, which differs for each setting. The scaling factor we choose across all setting is 1.5 (and round up to the next nice whole number).

In the case of a supernova setting, we consider a distinct expert feature size of 6. This is because the maximum number of distinct expert features we can obtain is 6, given that there are a maximum of 3 humps in the time series dataset. For each hump, there are both peaks and troughs, leading to a potential maximum of 6 distinct expert features.

For the multilingual politeness setting, the group maximum would be 40, which is the total number of lexical categories, 26, with the scaling factor multiplied in to give some flexibility. 

For the emotion setting, the group maximum would be , which is the total number of lexical categories, 26, with the scaling factor multiplied in to give some flexibility. 

For mass maps, the group maximum would be 25. We compute the maximum number of local maximums 7 on mass maps blurred with $\sigma=3$ and local minimums 7 on mass maps blurred with $\sigma=5$, which sums up to be 14. We can then multiply with the scaling factor to give some flexibility and then we round up to 25.

\paragraph{Baseline Parameters.}
For mass maps, we use the following parameters for baselines. For patch, we use $8\times 8$ grid. For QuickShift, we use kernel size 5, max dist 10, and sigma 0.2. For watershed, we use min dist 10, compactness 0. For SAM, we use `vit\_h'. For Archipelago, we use the same Quickshift parameters for the Quickshift segmenter.

\paragraph{Baseline Results.}
We report the full baseline results with standard deviations in Table~\ref{tab:baseline_scores_full}.

\input{figures/baseline_scores}

%% file: figures/baseline_scores.tex
\begin{table}[t]
\small
\centering
\begin{tabular}{llcccccc}
\toprule
   & \textbf{Method} 
    & \textbf{Cholecystectomy}
    & \textbf{Chest X-ray}
    & \textbf{Mass Maps} \\
    
    \cmidrule(lr){2-2} \cmidrule(lr){3-5}
    \multirow{7}{*}{\textit{Image }} &Identity
       & 0.4648 $\pm$ 0.0045
       & 0.2154 $\pm$ 0.0027
       & 0.5483 $\pm$ 0.0015
       \\
    &Random
        & 0.1084 $\pm$ 0.0004
        & 0.0427 $\pm$ 0.0001
        & 0.5505 $\pm$ 0.0014
        \\
    &Patch 
        & 0.0327 $\pm$ 0.0001
        & 0.0999 $\pm$ 0.0008
        & 0.5555 $\pm$ 0.0013 \\
    &Quickshift
        & 0.2664 $\pm$ 0.0036
        & 0.3419 $\pm$ 0.0025
        & 0.5492 $\pm$ 0.0009 \\
    &Watershed
        & 0.2806 $\pm$ 0.0049
        & 0.1452 $\pm$ 0.0017
        & 0.5590 $\pm$ 0.0017 \\
    &SAM
        & 0.3642 $\pm$ 0.0092
        & 0.3151 $\pm$ 0.0064
        & 0.5521 $\pm$ 0.0009
        \\
    &CRAFT
        & 0.0278 $\pm$ 0.0003
        & 0.1175 $\pm$ 0.0011
        & 0.3996 $\pm$ 0.0023
        \\
    \cmidrule(lr){1-5}
    \multirow{2}{*}{\textit{Domain-Agnostic}} &Clustering
        & 0.2839 $\pm$ 0.0024
        & 0.2627 $\pm$ 0.0039
        & 0.5515 $\pm$ 0.0014
        \\
    &Archipelago
        & 0.3271 $\pm$ 0.0076
        & 0.2148 $\pm$ 0.0009
        & 0.5542 $\pm$ 0.0014
        \\
\midrule
    & & \textbf{Supernova} \\
    \cmidrule(lr){3-5}
    \multirow{5}{*}{\textit{Time Series}} &Identity & 0.0152 $\pm$ 0.0011
       \\
    &Random & 0.0358 $\pm$ 0.0021
        \\
    &Slice 5
        & 0.0337 $\pm$ 0.0015 \\
    &Slice 10
        & 0.0555 $\pm$ 0.0044 \\
    &Slice 15
        & 0.0554 $\pm$ 0.0032 \\
    \cmidrule(lr){1-5}
    \multirow{2}{*}{\textit{Domain-Agnostic}} &Clustering
        & 0.2622 $\pm$ 0.0037
        \\
    &Archipelago
        & 0.2574 $\pm$ 0.0082
        \\
\midrule
    & & \textbf{Multilingual Politeness} & \textbf{Emotion} \\
    \cmidrule(lr){3-5}
   \multirow{5}{*}{\textit{Text}} & Identity
       & 0.6070 $\pm$ 0.0015
       & 0.0103 $\pm$ 0.0001
    \\
    & Random
        & 0.6478 $\pm$ 0.0012
        & 0.0303 $\pm$ 0.0004
        \\
    & Words
        & 0.6851 $\pm$ 0.0010 
        & 0.1182 $\pm$ 0.0003
        \\
    & Phrases
        & 0.6351 $\pm$ 0.0010
        & 0.0198 $\pm$ 0.0003
        \\
    & Sentences
        & 0.6109 $\pm$ 0.0006
        & 0.0120 $\pm$ 0.0002
        \\
    \cmidrule(lr){1-5}
   \multirow{2}{*}{\textit{Domain-Agnostic}} & Clustering 
        & 0.6680 $\pm$ 0.0048
        & 0.0912 $\pm$ 0.0005
        \\
    & Archipelago
        & 0.6773 $\pm$ 0.0006 
        & 0.0527 $\pm$ 0.0008
        \\
\bottomrule
\end{tabular}

\vspace{0.25cm}
\caption{Baselines of different \dataset settings. We report the mean \datasetscore for all examples in each setting, with standard deviations.
}
\label{tab:baseline_scores_full}
\end{table}



%% file: appendix/dataset_example_features/cosmology.tex
\subsection{Mass Maps Dataset}
\label{app:mass_maps_example_features}
\paragraph{Example Features.} 

\input{figures/massmaps_quickshift}

As MassMaps does not have annotated expert features, we only show example of generated features with corresponding percent void and cluster and alignment scores in Figure~\ref{fig:massmaps_quickshift}.
We can see that the 6th feature (3rd image on the second row) achieves the highest alignment score with a large percentage of void (86.3\%) and a very small percent of cluster (0.8\%), while the 5th features (2nd image on the second row) has the lowest alignment of (57.3\%), as it is not fully aligned to either void or cluster.


%% file: figures/massmaps_quickshift.tex
\begin{figure}[t]

\centering

Quickshift Features

\includegraphics[width=0.8\linewidth]{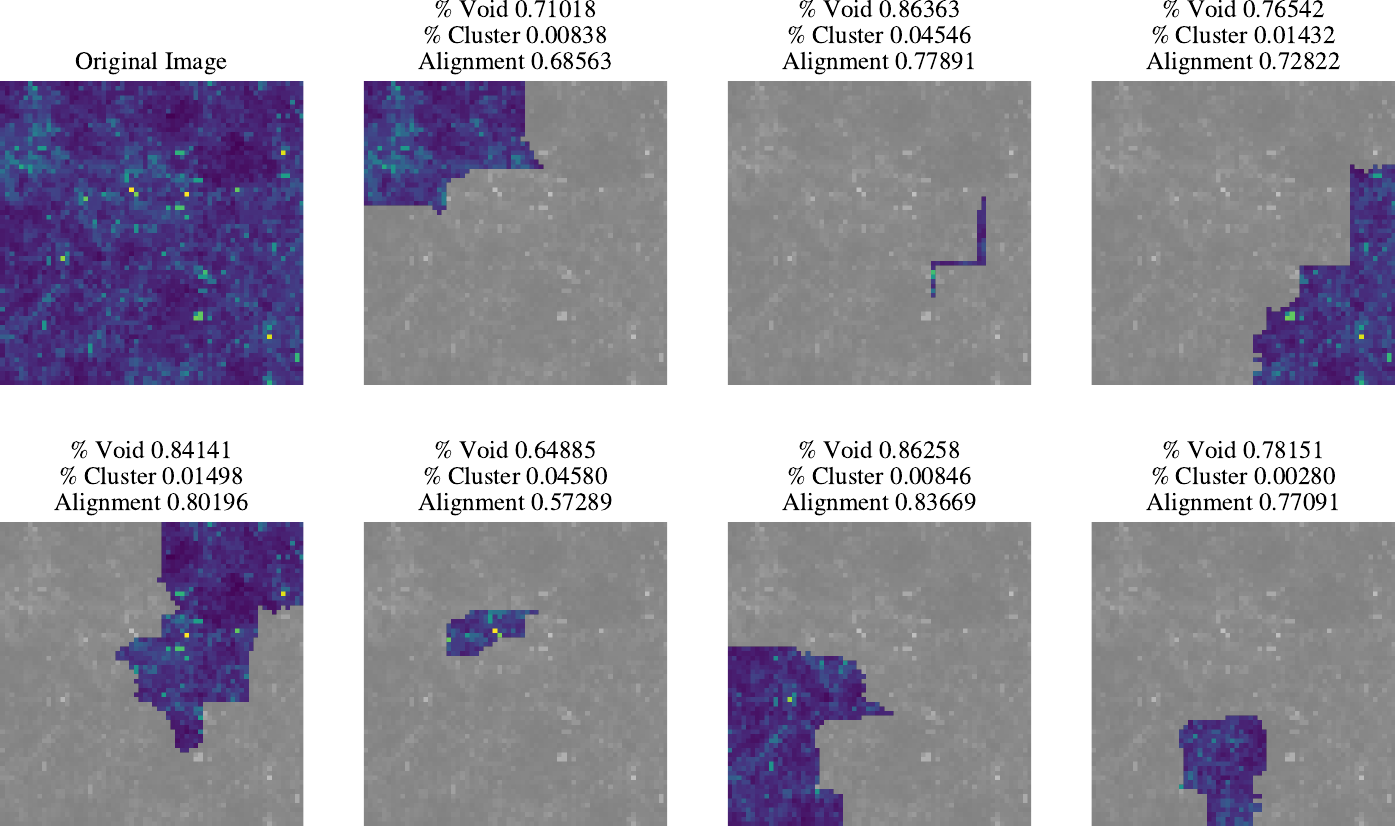}

\vspace{10pt}

\caption{MassMaps features from quickshift with void, cluster, and expert alignment scores.}
\label{fig:massmaps_quickshift}
\end{figure}

%% file: appendix/dataset_example_features/supernova.tex
\subsection{Supernova Dataset}
\label{app:supernova_example_features}

\input{figures/supernova_clustering}

See Figure~\ref{fig:supernova_clustering}.

%% file: figures/supernova_clustering.tex
\begin{figure}[t]
    \centering

    \begin{minipage}{0.48\textwidth}
        \centering
        \includegraphics[width=\linewidth]{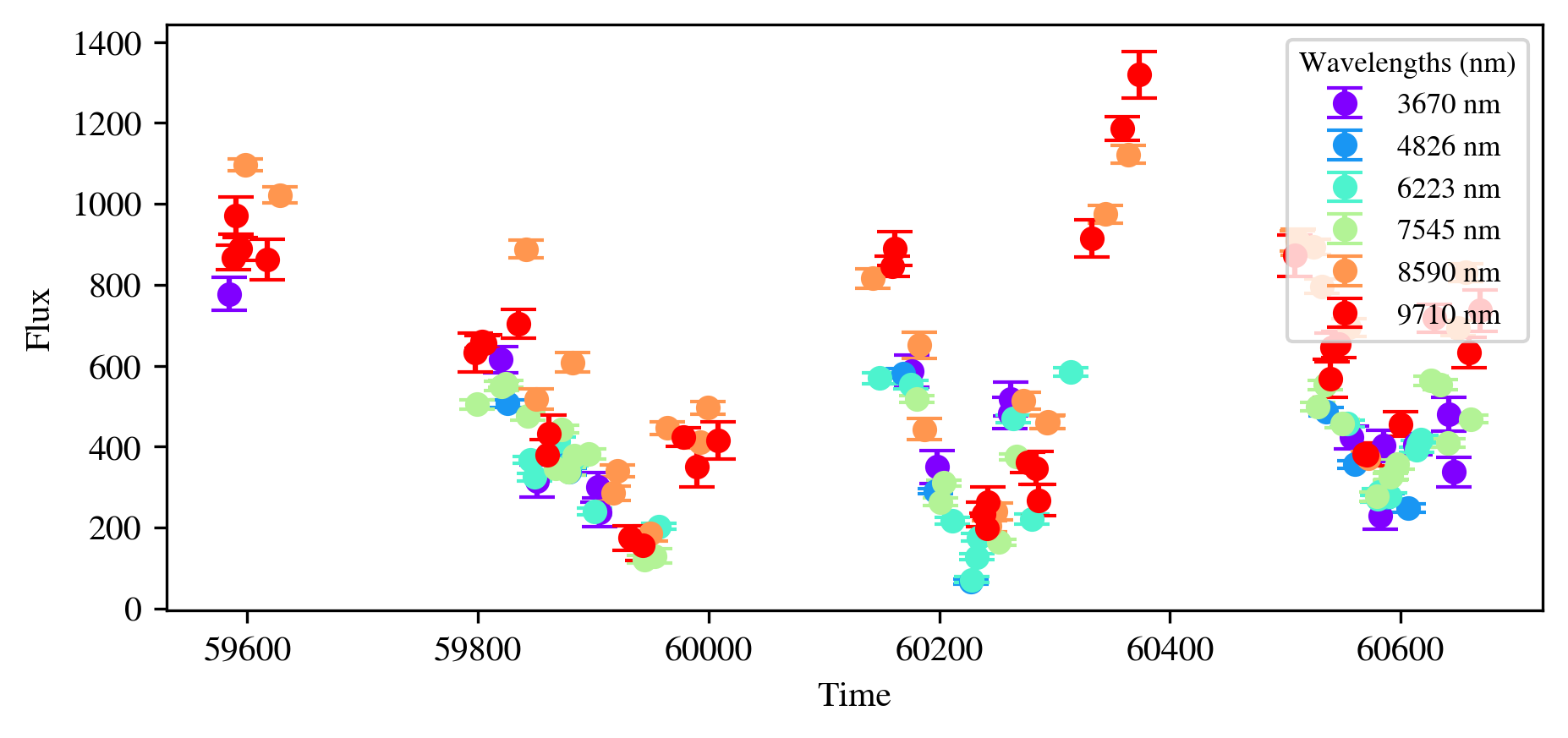}
        \par\smallskip
        \small Original Image
    \end{minipage}\hfill     \begin{minipage}{0.48\textwidth}
        \centering
        \includegraphics[width=\linewidth]{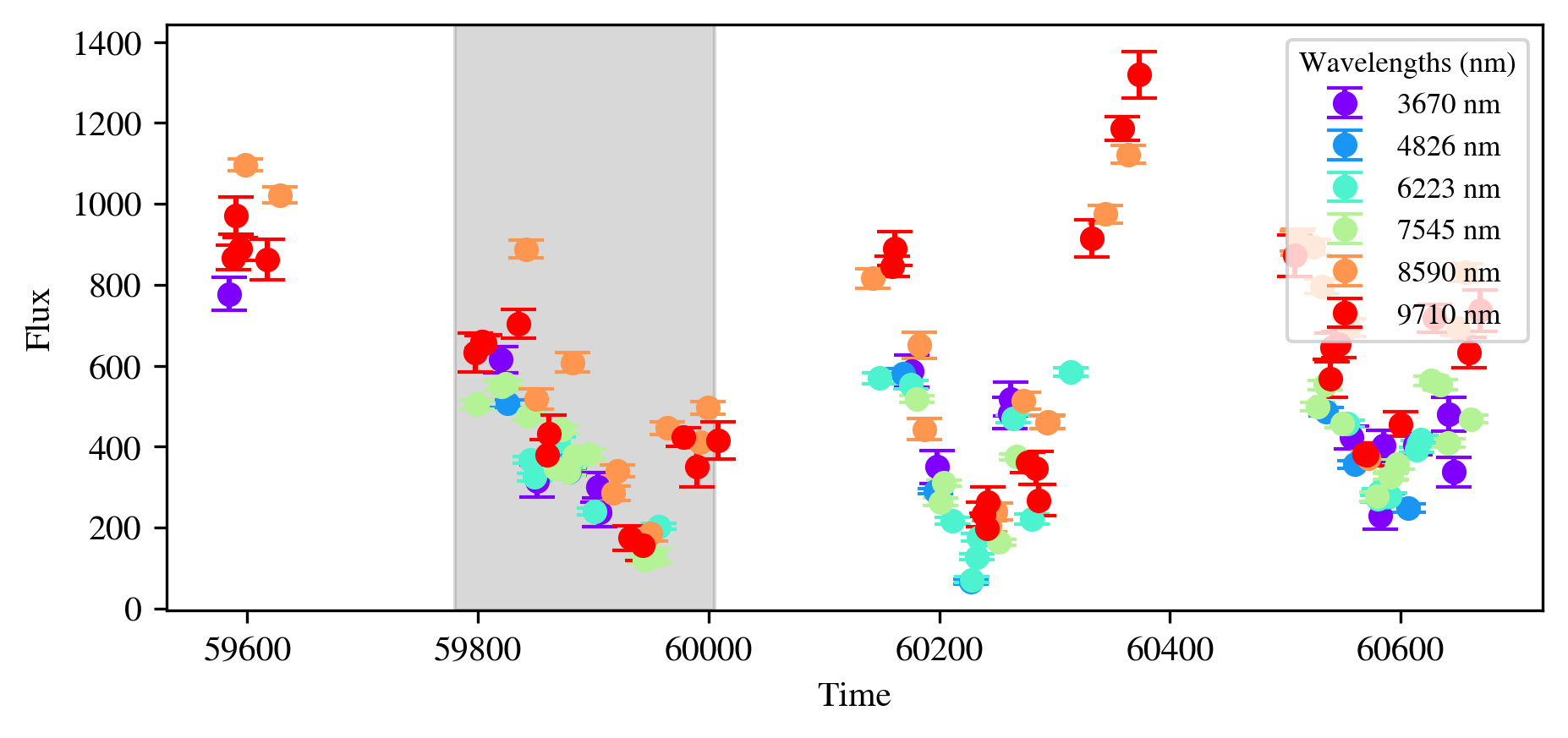}
        \par\smallskip
        \small Clustering Feature1
    \end{minipage}
    
    \vspace{1em}
    \begin{minipage}{0.48\textwidth}
        \centering
        \includegraphics[width=\linewidth]{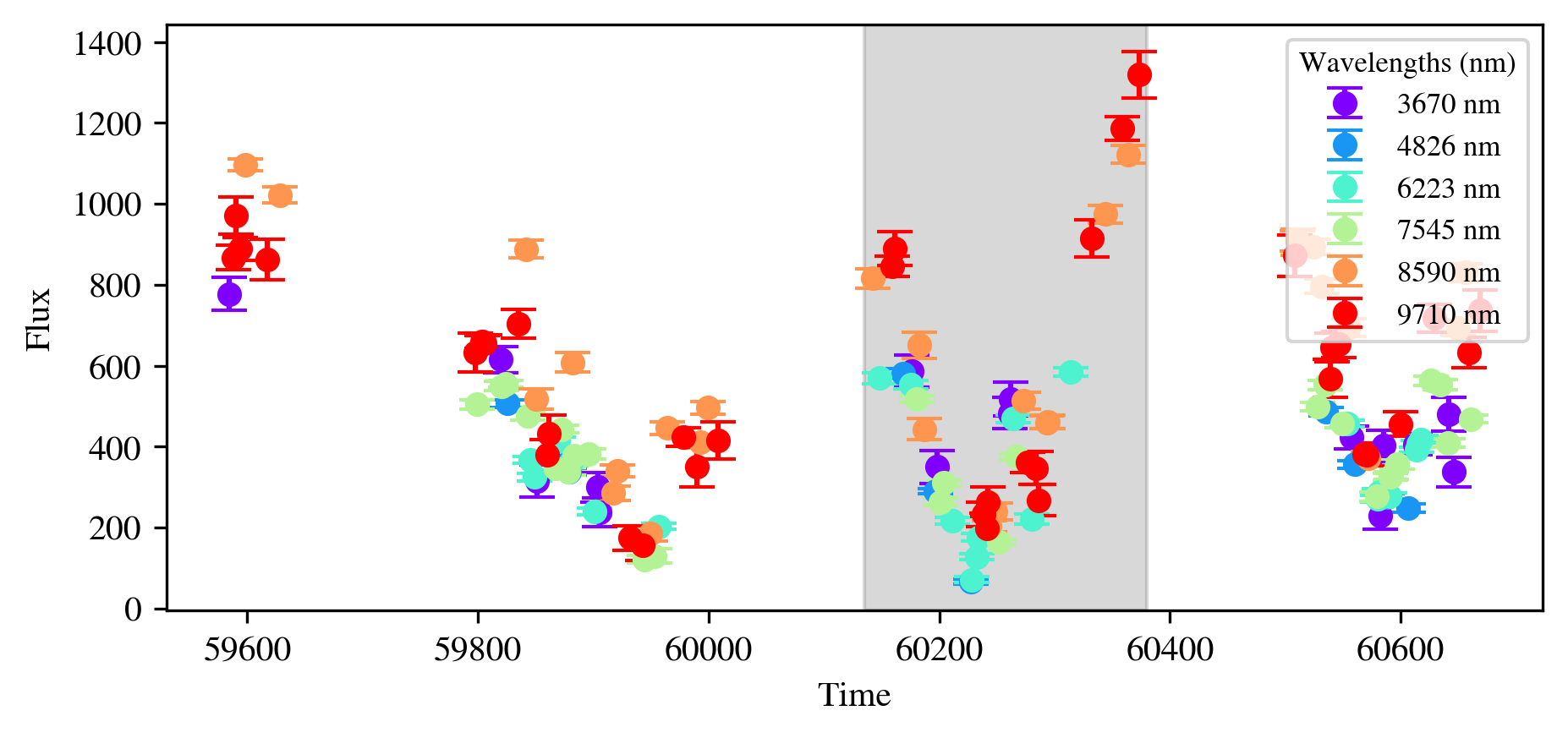}
        \par\smallskip
        \small Clustering Feature2
    \end{minipage}\hfill
    \begin{minipage}{0.48\textwidth}
        \centering
        \includegraphics[width=\linewidth]{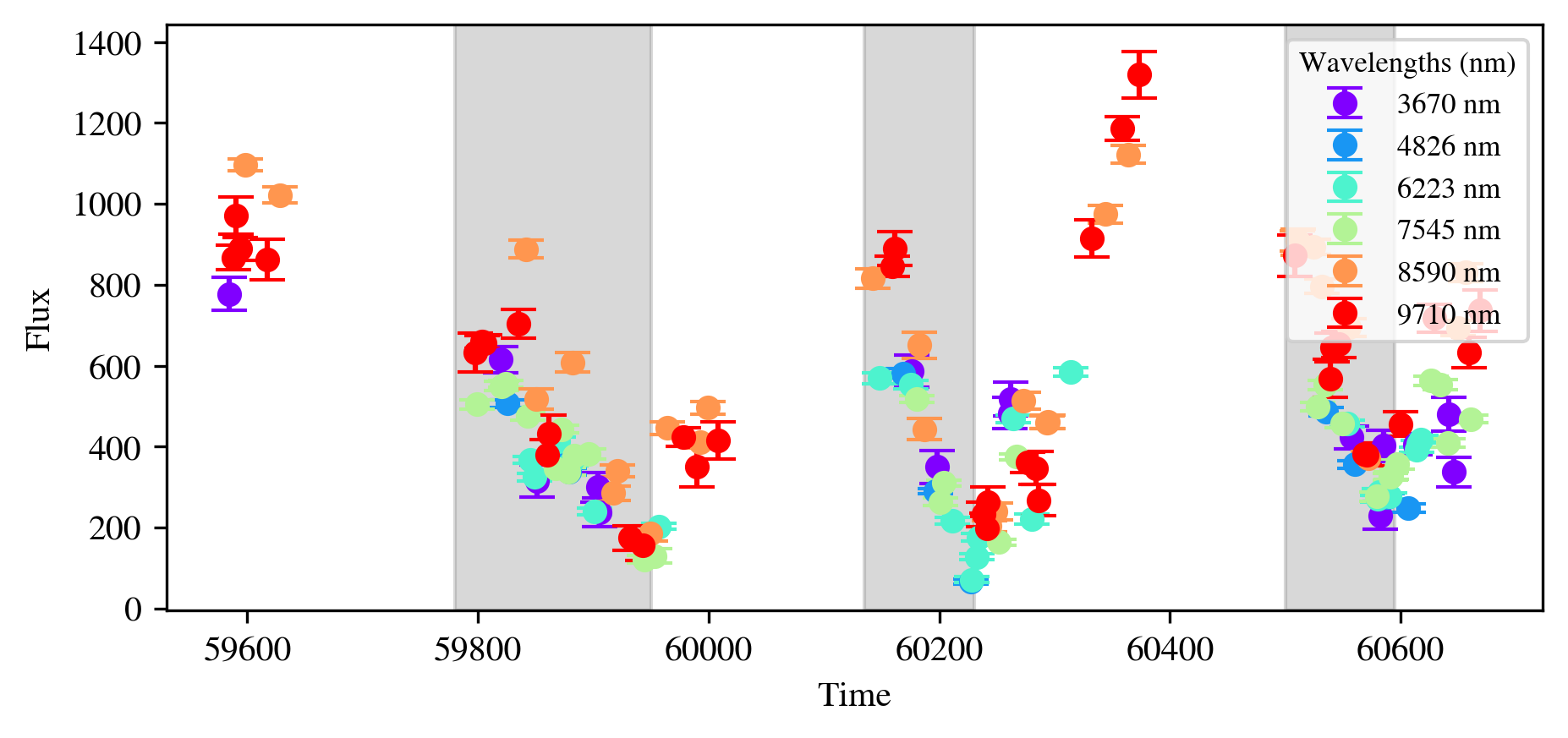}
        \par\smallskip
        \small Clustering Feature3
    \end{minipage}

    \caption{Supernova features from clustering.}
    \label{fig:supernova_clustering}
\end{figure}

%% file: appendix/dataset_example_features/politeness.tex
\subsection{Multilingual Politeness Dataset}
\label{app:politeness_example_features}
\paragraph{Example Features.} Since the multilingual politeness dataset does not have annotated expert features, we use semantic similarity with the politeness lexica in \citet{havaldar-etal-2023-comparing}, adapted from the Stanford Politeness Lexicon \citep{danescu2013computational}.

A feature for the multilingual politeness dataset is a single word. We choose to not further break down words into tokens, as it is unclear what the cosine similarity between a token and a word in a lexicon would mean. In this vein, feature groups are a collection of words in the input that need not appear consecutively.

\paragraph{Expert Features.} An expert feature is a lexical category from the Stanford Politeness Lexicon \citep{danescu2013computational}. Such categories include apology words, greetings, positive sentiment words, etc., where each category is either an indicator of politeness or an indicator of rudeness. see Table~\ref{tab:politeness-examples} for examples of such expert features.

\begin{table*}[t]
    \centering
    \small
    \begin{tabular}{llp{4cm}}
    \toprule
     \textbf{Input} & \textbf{Example Feature} & \textbf{Expert Feature} \\
     \midrule
     \multirow{6}{6cm}{I was running my spellchecker and totally didn't realize that this was a vandalized page. Please accept my apology. I will spellcheck a little slower next time.} & ``my'' & First-person pronouns: \textit{I, my, mine, etc.}\\
     & ``vandalized'' & Negative sentiment: \textit{bad, ugly, terrorized, etc.} \\
     & ``apology'' & Apologizing: \textit{sorry, apology, my bad, etc.} \\
     \bottomrule
    \end{tabular}
    \caption{Example features and corresponding expert features for the multilingual politeness dataset.}
    \label{tab:politeness-examples}
\end{table*}

%% file: appendix/dataset_example_features/emotion.tex
\subsection{Emotion Dataset}
\label{app:emotion_example_features}
\paragraph{Example Features.} The emotion dataset also does not have annotated expert features, so we use valence and arousal signal \citep{russell1980circumplex}.

A feature for the emotion dataset is a single word. We choose to not further break down the words into tokens, as it is unclear what the projection of a single token onto the valence-arousal plane would mean. A group is a collection of words in the input that need not appear consecutively.  

\paragraph{Expert Features.} An expert feature is a word that is extremely close to an axis point on the valence arousal plane - see Table~\ref{tab:axis_definitions} or Table~\ref{tab:emotion-examples} for examples of such expert features.

\begin{table*}[t]
    \centering
    \small
    \begin{tabular}{llp{4cm}}
    \toprule
     \textbf{Input} & \textbf{Example Feature} & \textbf{Expert Feature} \\
     \midrule
     \multirow{6}{6cm}{This was potentially the most dangerous stunt I have ever seen someone do. One minor mistake and you die.} & ``dangerous'' & Low Valence: \textit{death, horrible, scary, etc.}\\
     & ``minor'' & Low Arousal: \textit{calm, tired, unexciting, etc.} \\
     & ``stunt'' & High Arousal: \textit{furious, excited, surprised, etc.} \\
     \bottomrule
    \end{tabular}
    \caption{Example features and corresponding expert features for the emotion dataset.}
    \label{tab:emotion-examples}
\end{table*}

%% file: appendix/dataset_example_features/chest_xray.tex
\subsection{Chest X-Ray Dataset}
\label{app:chestxray_example_features}

\input{figures/chestx_expert_vs_quickshift}

See Figure~\ref{fig:chestx_expert_vs_quickshift}.



%% file: figures/chestx_expert_vs_quickshift.tex
\begin{figure}[t]

\centering


Expert Features

\includegraphics[width=0.8\linewidth]{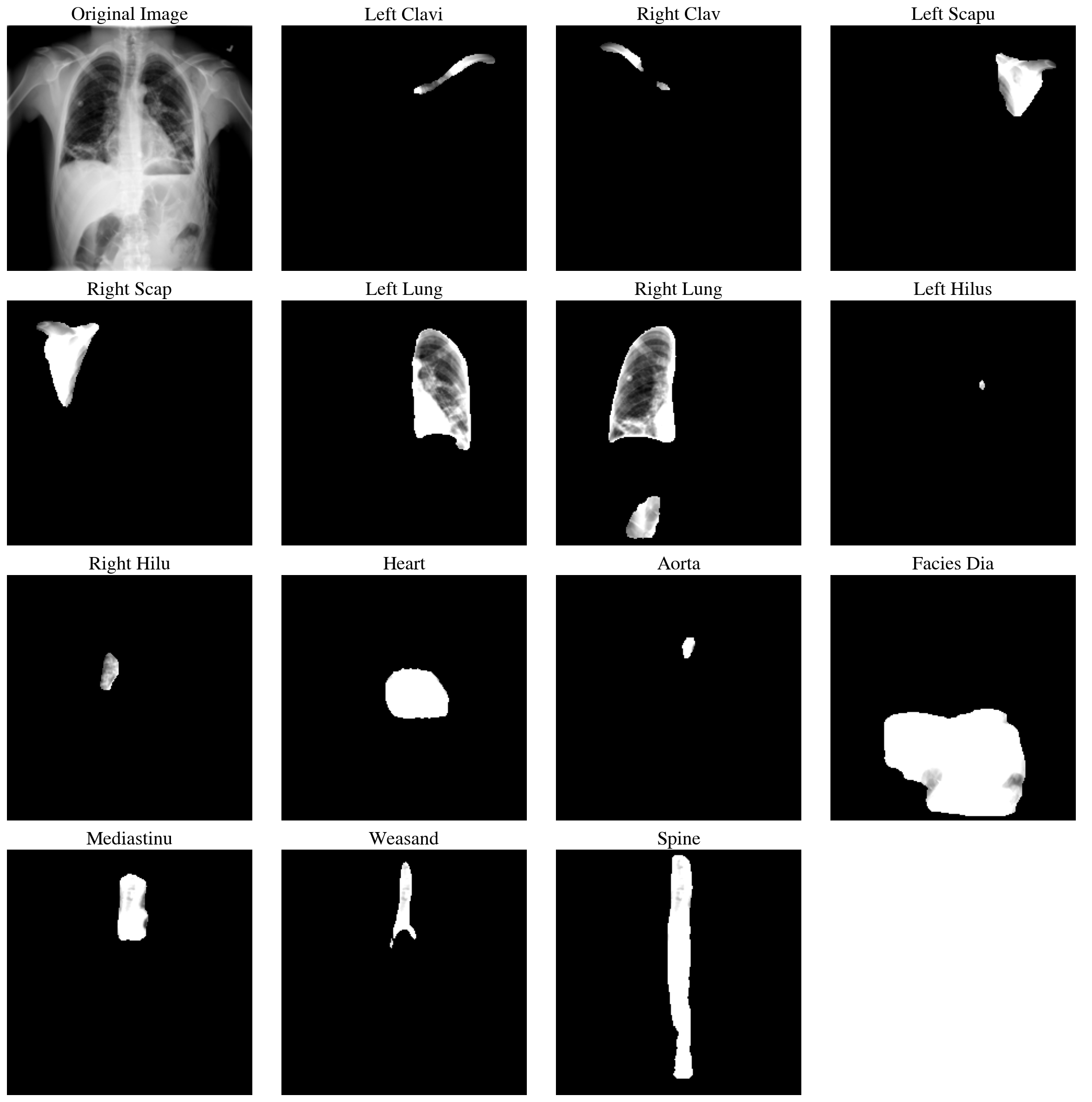}

\vspace{0.5cm}

Quickshift Features

\includegraphics[width=0.8\linewidth]{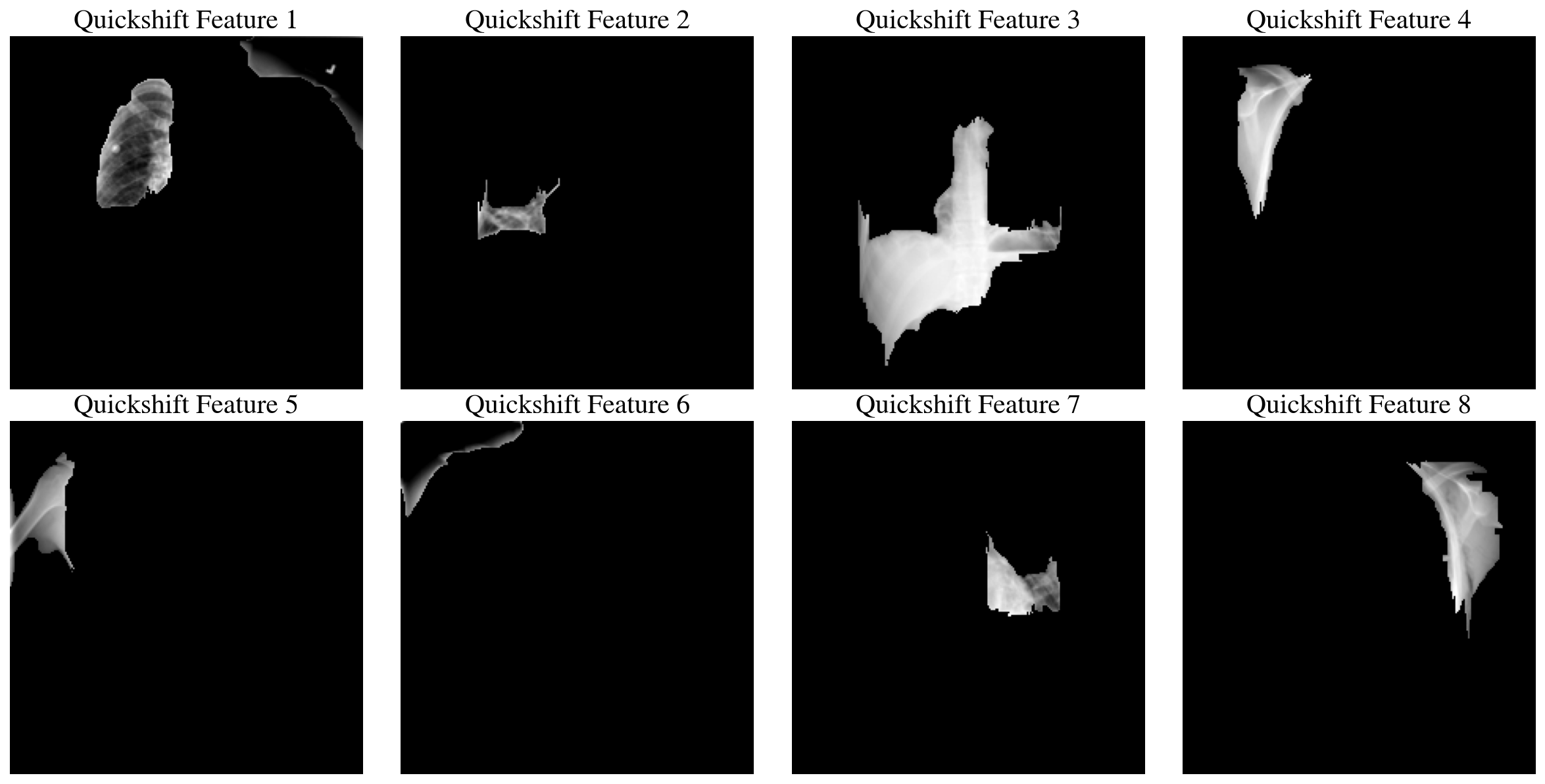}

\vspace{10pt}

\caption{Chest X-ray features from experts (top) and some samples from quickshift (bottom).}
\label{fig:chestx_expert_vs_quickshift}
\end{figure}

%% file: appendix/dataset_example_features/abdomen.tex
\subsection{Laparoscopic Cholecystectomy Surgery Dataset}
\label{app:abdomen_example_features}

\input{figures/cholec_expert_vs_quickshift}

See Figure~\ref{fig:cholec_expert_vs_quickshift}.



%% file: figures/cholec_expert_vs_quickshift.tex
\begin{figure}[t]

\centering


Expert Features

\includegraphics[width=0.9\linewidth]{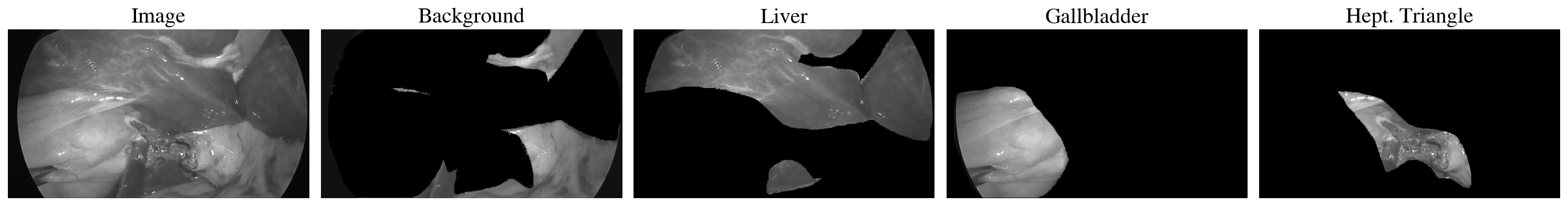}

\vspace{1cm}

Quickshift Features

\includegraphics[width=0.9\linewidth]{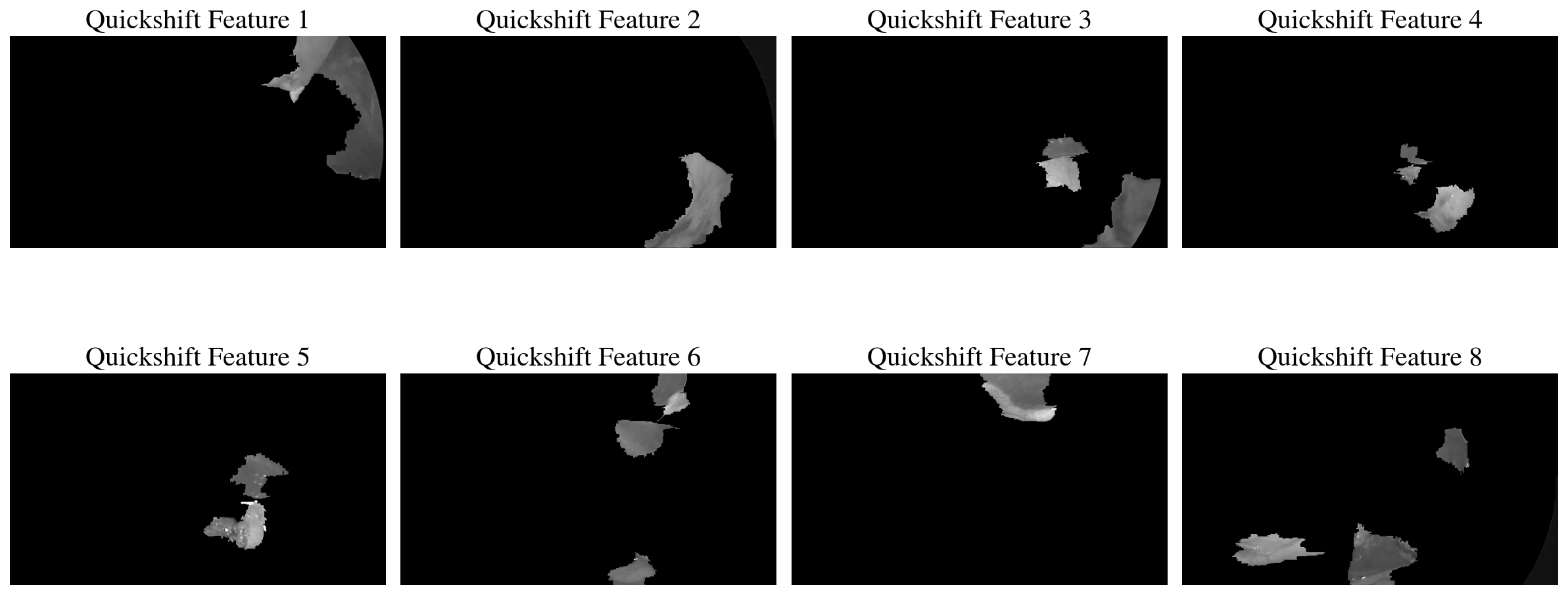}

\vspace{10pt}

\caption{Laparoscopic Cholecystectomy features from experts (top) and some samples from quickshift (bottom).}
\label{fig:cholec_expert_vs_quickshift}
\end{figure}

%% file: appendix/add_setting.tex
Here, we provide a step-by-step walkthrough for adding a new setting to the \dataset benchmark, so that the process may be more accessible to future researchers.  

\begin{enumerate}
    \item Determine if the new setting has explicit or implicit expert alignment.
    \item If the setting has explicit expert alignment, i.e. there are explicit annotations for expert features available, one can use the explicit's case's $\mrm{\expertalign}$ function, as shown in Equation \ref{eqn:expert_align}.
    \item Otherwise, if the setting has implicit expert alignment, one must define a custom expert alignment scoring function for that setting. \\
    \textit{Note:} We suggest consulting with experts of that domain so that the criteria incorporated in the formulation of the scoring function aligns well with expert judgment. 
    \item Once the expert alignment scoring function is defined, we can plug this into the \dataset framework, as defined in Equations \ref{eqn:featurealign} and \ref{eqn:datasetscore}, to obtain the \datasetscore for the setting.
    \item Depending on the data modality of the setting, one can run relevant baseline methods, including those we provide in Section \ref{baselines_section}.
\end{enumerate}

%% file: appendix/compute.tex
All experiments were conducted on two server machines, each with 8 NVIDIA A100 GPUs and 8 NVIDIA A6000 GPUs, respectively.

%% file: appendix/safeguards.tex
The datasets and models that we use in this work are not high risk and are previously open-source and publicly available. In particular, for our medical settings which would pose the most potential safety concern, the datasets we sourced our \dataset datasets from are already open-source and consists of de-anonymized images.

%% file: appendix/datasheet.tex
We follow the documentation framework provided by \citet{gebru2021datasheets} to create datasheets for the \dataset datasets. We address each section per dataset.

\subsection{Motivation}
\paragraph{For what purpose was the dataset created?}
\begin{itemize}[leftmargin=16pt,topsep=1pt,noitemsep]
    \item \textbf{Mass Maps}: The original dataset, CosmoGridV1~\citep{cosmogrid1}, was created to help predict the initial states of the universe in cosmology.
    \item \textbf{Supernova}: The original dataset PLAsTiCC for Kaggle competition~\citep{allam2018photometric}, was created to classify astronomical sources that vary with time into different classes. 
    \item \textbf{Multilingual Politeness}: The Multilingual Politeness dataset~\citep{havaldar-etal-2023-comparing} was created to holistically explore how politeness varies across different languages.
    \item \textbf{Emotion}: The original dataset, GoEmotions~\citep{demszky-etal-2020-goemotions}, was created to help understand emotion expressed in language.
    \item \textbf{Chest X-Ray}: The NIH-Google dataset~\citep{majkowska2020chest}, which is a relabeling of the NIH ChestX-ray14 dataset~\citep{wang2017chestx}, was created to help identify the presence of common pathologies.
    \item \textbf{Laparoscopic Cholecystectomy Surgery}: The original datasets from M2CAI16 workflow challenge~\citep{stauder2016tum} and Cholec80~\citep{twinanda2016endonet} were created to help identify the safe and unsafe areas of surgery.
\end{itemize}

\paragraph{Who created the dataset (e.g., which team, research group) and on behalf of which entity (e.g., company, institution, organization)?}
\begin{itemize}[leftmargin=16pt,topsep=1pt,noitemsep]
    \item \textbf{Mass Maps}: The original dataset CosmoGridV1~\citep{cosmogrid1} was created by Janis Fluri, Tomasz Kacprzak, Aurel Schneider, Alexandre Refregier, and Joachim Stadel at the ETH Zurich and the University of Zurich.
    The simulations were run at the Swiss Supercomputing Center (CSCS) as part of the project ``Measuring Dark Energy with Deep Learning'', hosted at ETH Zurich by the IT Services Group of the Department of Physics.
    We adapt the dataset and add a validation split.
    \item \textbf{Supernova}: The original dataset PLAsTiCC was created by \citet{theplasticcteam2018photometric}. We adapt the dataset, add a validation split, and balance the sets for each class.
    
    \item \textbf{Multilingual Politeness}: The Multilingual Politeness dataset~\citep{havaldar-etal-2023-comparing} was created by Shreya Havaldar, Matthew Pressimone, Eric Wong, and Lyle Ungar at the University of Pennsylvania.
    \item \textbf{Emotion}: The original GoEmotions~\citep{demszky-etal-2020-goemotions} dataset was created by Dorottya Demszky, Dana Movshovitz-Attias, Jeongwoo Ko, Alan Cowen, Gaurav Nemade, and Sujith Ravi at Stanford University, Google Research and Amazon Alexa.
    \item \textbf{Chest X-Ray}: The NIH-Google dataset~\citep{majkowska2020chest} was created by Anna Majkowska, Sid Mittal, David F Steiner, Joshua J Reicher, Scott Mayer McKinney, Gavin E Duggan, Krish Eswaran, Po-Hsuan Cameron Chen, Yun Liu, Sreenivasa Raju Kalidindi, et al., at Google Health, Stanford Healthcare and Palo Alto Veterans Affairs, Apollo Radiology International, and California Advanced Imaging.
    \item \textbf{Laparoscopic Cholecystectomy Surgery}: The M2CA116 workflow challenge dataset~\citep{stauder2016tum} was created by Ralf Stauder, Daniel Ostler, Michael Kranzfelder, Sebastian Koller, Hubertus Feußner, and Nassir Navab at Technische Universität München in Germany and Johns Hopkins University. The Cholec80 dataset~\citep{twinanda2016endonet} was created by Andru P Twinanda, Sherif Shehata, Didier Mutter, Jacques Marescaux, Michel De Mathelin, and Nicolas Padoy, at ICube, University of Strasbourg, CNRS, IHU, University Hospital of Strasbourg, IRCAD and IHU Strasbourg, France.
\end{itemize}

\paragraph{Who funded the creation of the dataset?}
\begin{itemize}[leftmargin=16pt,topsep=1pt,noitemsep]
    \item Please refer to each setting's respective papers for funding details.
\end{itemize}

\subsection{Composition}
\begin{itemize}[leftmargin=16pt,topsep=1pt,noitemsep]
    \item The answers are described in our paper. Please refer to Section \ref{sec:datasets} and Appendix \ref{app:dataset_details} for more details.
\end{itemize}

\subsection{Collection Process}
\begin{itemize}[leftmargin=16pt,topsep=1pt,noitemsep]
    \item We defer the collection process to the relevant works that created them. Please refer to Section \ref{sec:datasets} and Appendix \ref{app:dataset_details} for more details.
\end{itemize}

\subsection{Preprocessing/cleaning/labeling}
\begin{itemize}[leftmargin=16pt,topsep=1pt,noitemsep]
    \item The answers are described in our paper. Please refer to Section \ref{sec:datasets} and Appendix \ref{app:dataset_details} for more details.
\end{itemize}

\subsection{Uses}
\begin{itemize}[leftmargin=16pt,topsep=1pt,noitemsep]
    \item The answers are described in our paper. Please refer to Section \ref{sec:datasets} and Appendix \ref{app:dataset_details} for more details.
\end{itemize}

\subsection{Distribution}
\paragraph{Will the dataset be distributed to third parties outside of the entity (e.g., company, institution,
organization) on behalf of which the dataset was created?}
\begin{itemize}[leftmargin=16pt,topsep=1pt,noitemsep]
    \item No. Our datasets will be managed and maintained by our research group.
\end{itemize}

\paragraph{How will the dataset will be distributed (e.g., tarball on website, API, GitHub)?}
\begin{itemize}[leftmargin=16pt,topsep=1pt,noitemsep]
    \item The \dataset datasets are released to the public and hosted on Huggingface (please refer to links in Appendix \ref{dataset_links}).
\end{itemize}

\paragraph{When will the dataset be distributed?}
\begin{itemize}[leftmargin=16pt,topsep=1pt,noitemsep]
    \item The datasets have been released now, in 2024.
\end{itemize}

\paragraph{Will the dataset be distributed under a copyright or other intellectual property (IP) license, and/or under applicable terms of use (ToU)?}
\begin{itemize}[leftmargin=16pt,topsep=1pt,noitemsep]
    \item \textbf{Mass Maps}: The Mass Maps dataset is distributed under CC BY 4.0, following the original dataset CosmoGridV1~\citep{cosmogrid1}.
    \item \textbf{Supernova}: The Supernova dataset is distributed under the MIT license.
    \item \textbf{Multilingual Politeness}: The Multilingual Politeness dataset is distributed under the CC-BY-NC license.
    \item \textbf{Emotion}: The Emotion dataset is distributed under the Apache 2.0 license.
    \item \textbf{Chest X-Ray}: The Chest X-Ray dataset is distributed under the Apache 2.0 license.
    \item \textbf{Laparoscopic Cholecystectomy Surgery}: The Laparoscopic Cholecystectomy Surgery dataset is distributed under the CC by NC SA 4.0 license.
\end{itemize}


